\documentclass[letterpaper, 11 pt, onecolumn,oneside]{IEEEtran}
\usepackage{graphicx}
\usepackage{flushend}
\graphicspath{ {./images/} } 
\usepackage[boxruled,noline]{algorithm2e}
\usepackage{subcaption}
\usepackage{amsmath}
\usepackage{amssymb}
\usepackage{float}
\usepackage{amsfonts}
\usepackage{multirow}
\usepackage{tabularx}
\usepackage{algorithm2e}
\usepackage[colorlinks]{hyperref}
\usepackage[table]{xcolor} 
\usepackage{soul}
\usepackage[capitalise]{cleveref}

\DeclareMathOperator*{\CD}{cd_{mean}}
\DeclareMathOperator*{\CDhat}{\hat{cd}_{mean}}
\DeclareMathOperator*{\CDhatbar}{{\bar{\widehat{cd}_{mean}}}}
\DeclareMathOperator*{\TTC}{ttc_{min}}
\DeclareMathOperator*{\FD}{fd_{mean}}
\DeclareMathOperator*{\DR}{dr_{mean}}

\begin{document}

\title{Diagnosing and Predicting Autonomous Vehicle Operational Safety Using Multiple Simulation Modalities and a Virtual Environment}

\author{
\IEEEauthorblockN{Joe Beck\IEEEauthorrefmark{1}, Shean Huff \IEEEauthorrefmark{2}, Subhadeep Chakraborty \IEEEauthorrefmark{3} }\\

\IEEEauthorblockA{\IEEEauthorrefmark{1}\emph{jbeck9@vols.utk.edu} \IEEEauthorrefmark{2}\emph{contact.vilsolutions@gmail.com}  \IEEEauthorrefmark{3}\emph{schakrab@utk.edu}} \\

\IEEEauthorblockA{Mechanical, Aerospace and Biomedical Engineering Department, University of Tennessee, Knoxville, 1512 Middle Dr, Knoxville, TN 37996, USA} 
}

\maketitle

\begin{abstract}
Even as technology and performance gains are made in the sphere of automated driving, safety concerns remain. Vehicle simulation has long been seen as a tool to overcome the cost associated with a massive amount of on-road testing for development and discovery of safety critical "edge-cases". However, purely software-based vehicle models may leave a large realism gap between their real-world counterparts in terms of dynamic response, and highly realistic vehicle-in-the-loop (VIL) simulations that encapsulate a virtual world around a physical vehicle may still be quite expensive to produce and similarly time intensive as on-road testing. In this work, we demonstrate an AV simulation test bed that combines the realism of vehicle-in-the-loop (VIL) simulation with the ease of implementation of model-in-the-loop (MIL) simulation. The setup demonstrated in this work allows for response diagnosis for the VIL simulations. By observing causal links between virtual weather and lighting conditions that surround the virtual depiction of our vehicle, the vision-based perception model and controller of \textit{Openpilot}, and the dynamic response of our physical vehicle under test, we can draw conclusions regarding how the perceived environment contributed to vehicle response. Conversely, we also demonstrate response prediction for the MIL setup, where the need for a physical vehicle is not required to draw richer conclusions around the impact of environmental conditions on AV performance than could be obtained with VIL simulation alone. These combine for a simulation setup with accurate real-world implications for edge-case discovery that is both cost effective and time efficient to implement.

\hfill\break%
\noindent\textit{Keywords}: Connected and Automated Vehicle, Simulation, Hardware-in-the-loop Testing, Digital Twin, Virtual World

\end{abstract}

\section{Introduction}
The key research question that we attempt to address in this paper is how to systematically assess the safety of connected and automated vehicles (CAV) before driving them on public roads. The history of human-driven vehicle safety reveals a tendency to leap first and analyze later. Based on other concurrent research thrusts, it seems that there are mixed definitions for safe testing presently established, and the lack of an efficient safe method is both placing human lives at risk and threatening to stymie the development of this emerging technology. The goals of this paper are to introduce and outline a hybrid testing protocol that can enable future creation of a set of certification standard recommendations. We focus on level 3 automation in this paper as vehicles with this capability already exist in cars like Teslas and Cadillacs. These cars have level 3 capabilities but drivers are expected to monitor the systems to ensure proper operations. Given that several recent accidents have demonstrated that vision systems are prone to rare but catastrophic failures, the perception subsystem and the downstream effect of its inaccuracies on measurable driving metrics is an important diagnostic step in the evolution of AVs.  

With the increasing penetration of learning-based perception modules in real-world systems, it is imperative to better understand the inner workings of deep-learning models designed for computer-vision, and make them safe for widespread use. Traditionally, learning-enabled sensing/perception and control modules of AVs are designed and trained in isolation without system-level safety objectives. In this paper, we emphasize end-to-end safety of the overall driving system both during training and deployment. 

From the AV perspective, safety guarantees have been broadly pursued along two disparate paths: (i) \textit{robustness of perception systems} when subjected to adversarial or benign/natural perturbations, and (ii) \textit{robustness of control systems} that leverage reinforcement learning (RL). Adversarial robustness of image recognition systems have been investigated extensively over the past five years~\cite{carlini2017towards,madrytowards} under a variety of settings; more recently the focus has shifted towards natural robustness~\cite{miller2021accuracy,feuer2022meta}. A similar line of research can also be found for RL models, primarily with respect to adversarial perturbations to state/observation and action spaces, with different approaches pursuing various styles of robustness objectives~\cite{amani2021safe, dalal2018safe,srinivasan2020learning,Yang2023,thananjeyan2021recovery,emam2022safe, cheng2019endtoend}. However, the large majority of the above approaches use somewhat ad-hoc notions of safety/robustness, and do not explicitly address system-wide safety and especially concepts of safety aligned to human values. 


Another exciting direction for enusuring safety in robot behavior is the use of formal methods ~\cite{mindom2021assessing,Fulton_Platzer_2018,Alshiekh:2018:Safe}. 
Addressing concerns of predictability and safety in learned robot behaviors, \cite{li2020formal} proposed a formal methods approach to interpretable RL for robotic planning, leveraging formal methods and control theory. To ensure safety in the RL agent behaviors, the approach provides a formal specification language to explicitly define undesirable behaviors and uses control barrier functions (CBF) to enforce safety constraints. Similarly, in \cite{mindom2021assessing}, the authors present an approach that uses formal methods to assess the safety of an RL agent by designing moving adversaries that identify potential safety violations and implementing defense mechanisms to strengthen the agent's policy. Moreover, in \cite{Fulton_Platzer_2018} the authors present a technique that combines the exploration and optimization capabilities of learning with the safety guarantees of formal verification to ensure safe control through proof and learning. Overall, these papers highlight the importance of safety in RL for robotic planning and offer promising approaches to ensure the safe behaviors of RL agents.

Along with theoretical development in the field of safe and reliable learning enabled systems, AV research has been traditionally supported by large-scale testing programs on public roads, proving grounds (such as MCity) and to a large extent through the use of simulators. Even though there are no standard criteria to classify driving simulators, specially for AV research; traditionally one of the commonly accepted criteria is to classify the simulator based on fidelity and configurability \cite{eryilmaz2014novel}. The fidelity is determined by taking general characteristics under consideration such as visual, motion and auditory cues. In high-fidelity driving simulator systems, an up-to 360-degree field-of-view (FOV) is available with high-resolution projectors and complex computer-generated roadway scenes. Vehicle movements and sounds can be simulated with a varying level of realism. In a lower fidelity system, the driving simulator is usually a screen-based desktop simulator which has a limited FOV and the virtual vehicle is controlled via a set of steering wheel and pedals without a motion base. The National Advanced Driving Simulator (NADS-1) and NADS MiniSim at the University of Iowa are separate examples of high-fidelity and medium-fidelity driving simulators \cite{chen2001nads}. With increased research emphasis on SAE L3,L4 solutions, virtual simulators with no hardware component have also gained increasing popularity \cite{dosovitskiyCARLAOpenUrban2017b, rongLGSVLSimulatorHigh2020a, ramakrishnaANTICARLAAdversarialTesting2022}. Leveraging these virtual platforms for insights into AV behavior including accident analysis \cite{BeckAutomated2022}, automated edge-case discovery \cite{ramakrishnaANTICARLAAdversarialTesting2022}, and the development of formalized safety-based scenario definers like \textit{Scenic} \cite{FremontScenic} have given further credence to the idea that virtual worlds could be critical in the field of formal edge-case discovery for automated vehicles.

In this work, we demonstrate a safe systems testing protocol using a complete code-to-road (CTR) ecosystem, comprising of a real vehicle-in-the-loop simulator. We introduce the unique capabilities of the wheel speed synchronized steerable dynamometer facility developed at the University of Tennessee that enables full-sized vehicles to be immersed into a virtual replica of on-road scenarios, offering safe and repeatable technology evaluation. This unique platform lets us test the full operational envelope of the AVs, with the digital twin enabling the creation and evaluation of a wide range of situations with easily manageable safety consequences. The vehicle used for the experiments reported in this paper is a Level 3 SUV built and instrumented with a dedicated computer with ROS and CAN communication. The dynamometer can be programmed to mimic the road load. In this paper, we report the use of a real vehicle with hardware-in-the-loop processing, control and actuation, which lets us realistically test the effect of environmental and controller perturbations on self driving functionalities.

\begin{figure*}[t]
\centering
\includegraphics[width=\textwidth,]{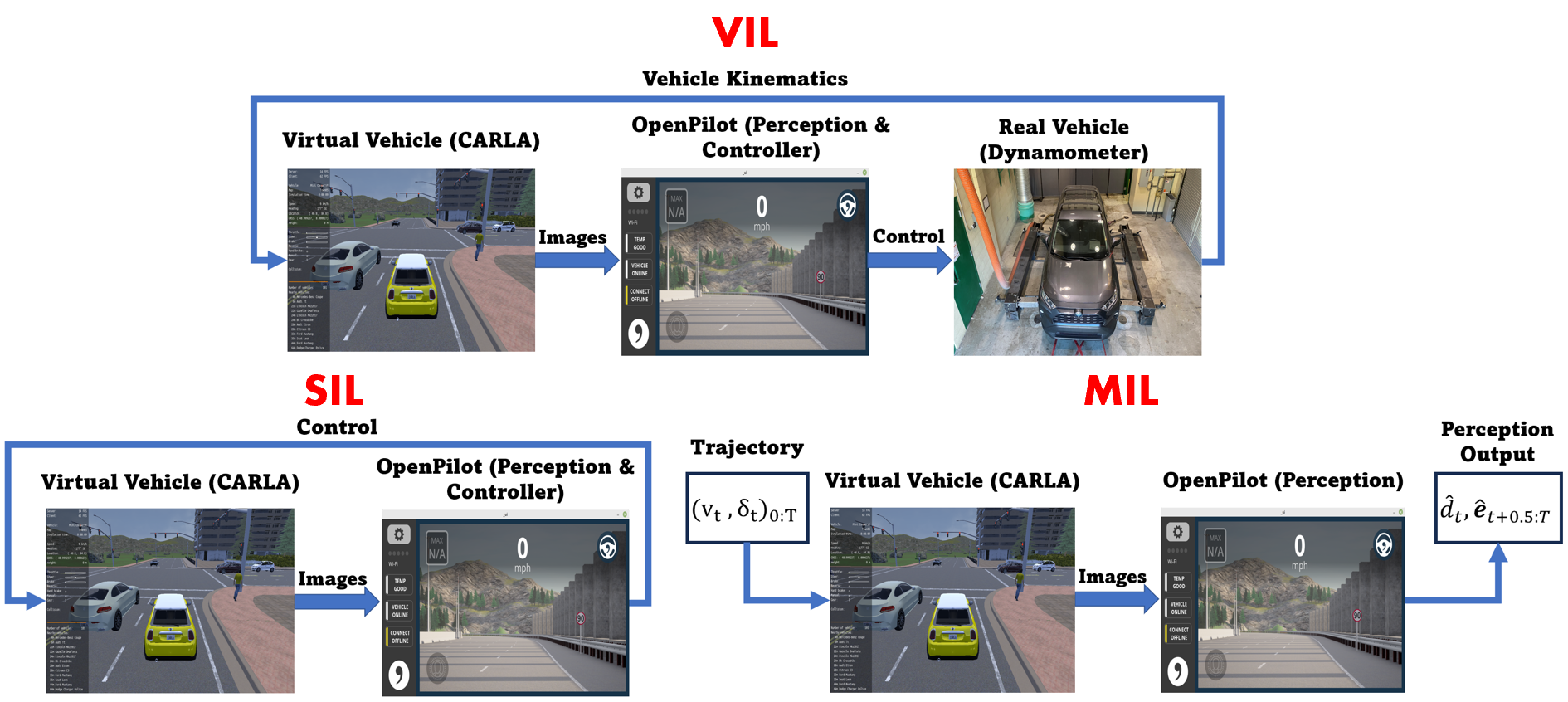}
\caption{An overview of the Vehicle-in-the-loop(VIL), Software-in-the-loop(SIL) and Model-in-the-loop(MIL) simulations used in this work. In VIL, the feedback to the simulator is vehicle kinematics measured from the CAN bus of the vehicle directly. In SIL, the controller feeds control commands to the simulator directly, and the physics model within CARLA determines the dynamic response. In MIL, the CARLA vehicle follows a predetermined trajectory, feeding images to the \textit{Openpilot} perception model.
}
\label{simov}
\centering
\end{figure*}

\section{Methodology}

In this section, our vehicle-in-the-loop (VIL), software-in-the-loop (SIL), and model-in-the-loop (MIL) simulation strategies are outlined. All three strategies provide a variety of challenges, as well as varying degrees of benefit. VIL simulation provides the most realistic vehicle response by far, in the sense that real vehicle dynamics are measured, and the control mechanism works through the actual CAN bus on the physical vehicle. However, even with our custom dynamometer setup, VIL is the most expensive and time consuming type of simulation; requiring a physical vehicle, a large experimental footprint, and testing that must be safety conscience and performed in real-time. SIL replaces the actual vehicle with a vehicle plant model. This is much easier to implement due to the lack of any hardware component, but is primarily done in this work to demonstrate the large gap that exists between VIL and SIL simulations in terms of dynamic response to identical stimuli. Finally MIL simulation as we define it is simply a controlled test of the perception model along a fixed trajectory. MIL is the easiest type of simulation to implement, and can even be performed faster-than-real-time (FTRT). Because MIL operates on a fixed trajectory, it is not possible to infer about vehicle response directly. However, VIL and MIL can be combined to get the benefits of both types of simulation, while mitigating against the challenges of each. Demonstrating this combination on the real SAE L2 controller \textit{Openpilot} constitutes the primary contribution of this work.

\subsection{Virtual World}
The vehicle simulations described in this work leverage the CARLA open-source simulation platform \cite{dosovitskiyCARLAOpenUrban2017b}. CARLA is a virtual simulation platform that provides tools for vehicle simulation, particularly autonomous systems. Within the scope of this work, we leverage CARLA as a testing environment in all three types of simulation. In VIL simulation, camera images from CARLA are generated and fed to the vehicle controller, the kinematic response is measured directly from the vehicle on the dynamometer, and this kinematic response is used to update the position of the vehicle in CARLA, completely bypassing CARLA's built in physics model with the dynamic response from the real vehicle. In SIL simulation, the controller takes in camera images, the vehicle physics model in CARLA takes throttle, braking, and steering commands from the controller, and the position of the vehicle is determined by the physics model within CARLA. Finally in MIL simulation, a kinematic trajectory for the vehicle across time is pre-defined. The vehicle in CARLA is simply moved along this trajectory as if on rails, the camera images from CARLA are sent to the vehicle perception system, and the output of the perception system is captured. These constitute the three types of simulations used in our experiments, and are outlined visually in \cref{simov}.

\subsection{Perception Model \& Controller}
A modified version of the open-source controller \textit{Openpilot} v8.13 by Comma.AI was used in this research \cite{comma.aiOpenpilotOpenSource}. \textit{Openpilot} is an SAE level 2 controller featuring compatibility with many recent vehicle makes and models. While \textit{Openpilot} is openly sourced, it uses a proprietary vision model dubbed \textit{Supercombo}. This deep learning-based vision model provides two key outputs used for control. First, the model takes in time-sequential camera images and produces a target lane trajectory for the car to follow. Model predictive control (MPC) and proportional–integral–derivative control (PID) use this trajectory to produce the steering, throttle, and braking commands that are sent to the vehicle in order to follow and maintain this trajectory. While the model produces a full trajectory across multiple time-steps for every image, we focus on the trajectory prediction of 500ms in the future, denoted as $\hat{\textbf{l}}_{t + 500ms}$. This value, in Cartesian world coordinates, can be considered the model's estimate of where the vehicle will be in 500ms, while $\textbf{l}_{t}$ is the current position of the vehicle. Second, the model produces a distance estimate between itself and the lead vehicle. We note that \textit{Openpilot} produces this estimate as a combination of relative speed, absolute speed, and distance estimates of the lead vehicle from \textit{Supercombo}. For clarity, we only report the final estimate for stopping distance that is given to the longitudinal MPC controller in \textit{Openpilot}, and denote this value as $\hat{d}_{t}$, where the actual distance from the lead vehicle and the ego vehicle at any given time $t$ is $d_{t}$.

\begin{figure*}[t]
\centering
\includegraphics[width=\textwidth]{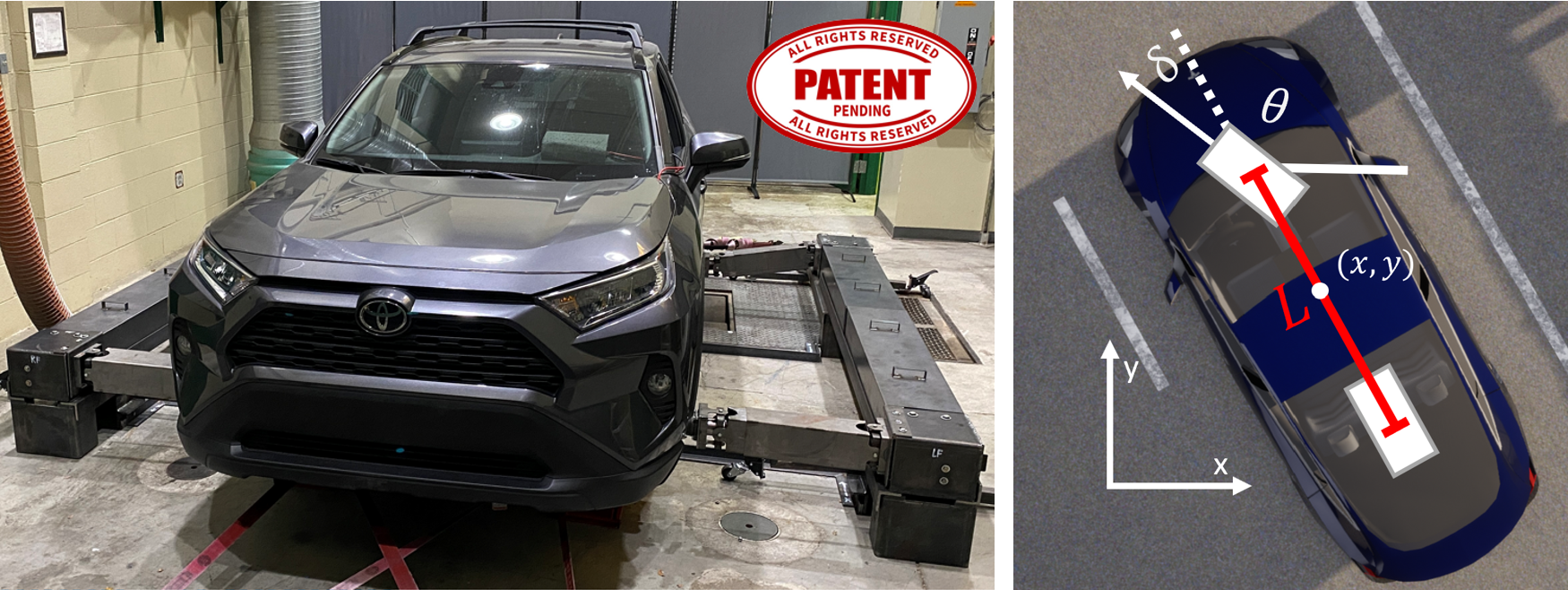}
\caption{The complete VIL Simulation setup is shown. (Left) The 2019 Toyota RAV4 is shown on the chassis dynamometer. The device that allows for steering and synchronized wheel speed is attached. (Right) The kinematic model used to update the position of the vehicle in CARLA using the dynamic response of the vehicle is shown. The wheelbase $L$, the heading $\theta$, and the steering angle $\delta$ are illustrated, the heading and the position of the vehicle $(x,y,\theta)$ are defined in terms of arbitrary world coordinates defined by the CARLA simulator.
}
\label{vilsetup}
\centering
\end{figure*}

\subsection{Vehicle-in-the-loop Simulation}
The vehicle used for VIL simulations was a stock 2019 Toyota RAV4, shown in \cref{vilsetup}. This vehicle was chosen primarily due to \textit{Openpilot}’s ability to integrate easily with the ADAS system in the vehicle, with no limitations on minimum cruise control speed or complete stop functionality. Communication between the vehicle and \textit{Openpilot} is performed through two vehicle CAN buses: one that interfaces with the on-board diagnostics (OBD II) port found on all modern vehicles in the United States, and another that interfaces with the Advanced driver-assistance systems (ADAS) CAN bus with a connector just above the rear view mirror. The \textit{Openpilot} controller typically works by using a windshield mounted camera combined with the vehicle’s built-in radar to perceive the road. In our setup, there is no physical camera facing the road. Instead, \textit{Openpilot} and CARLA are run simultaneously on a laptop, with images from CARLA being fed directly into \textit{Openpilot}, and control information being sent to the car from the computer via a USB-to-CAN connector provided by Comma.AI. Additionally, all radar and sensor-based localization functions within \textit{Openpilot} were disabled, making the controller’s perception system purely vision-based. 

All of the VIL simulations were done on a single-axle dynamometer equipped with an eddy-current brake for applying road load. VIL simulation is made possible on the chassis dynamometer by a unique prototype machine, provided by VIL Solutions, Inc., that is designed to deliver Steering-enabled Synchronized wheel-Speed (SeSwS) \footnote{The intellectual property associated with SeSwS is patent pending, owned exclusively by VIL Solutions Inc. Hardware has been provided for research purposes to the University of Tennessee, Knoxville in partnership with VIL Solutions Inc.} and shown in \cref{vilsetup}. This machine enables the ability to speed match all four wheels by linking the front and rear axles, and thus avoiding the typical challenges caused by vehicle safety system faults detected from wheel speed mismatches. This technology also allows for the actuation of the vehicle’s steering system through an articulated connection to the front axle. The combination of this equipment and the dynamometer allow for safe VIL testing in a controlled environment, where a virtual environment can provide stimulus to the vehicle.

For the VIL simulations, the dynamic response of the vehicle can be combined with the kinematic bicycle model to produce the next position of the vehicle within the simulation. The kinematic bicycle model is shown visually in \cref{vilsetup}. The four wheel speeds are measured from the CAN bus, averaged together and fed into a one-dimensional Kalman filter to produce an estimate of the total vehicle speed $\textbf{v}_{t}$. The equivalent steering angle of the ``bicycle wheel'' is computed as $\delta_w = \delta_{s} / sr$, where the steering ratio is $sr=14.3$ for the 2019 Toyota RAV4, and the orientation of the steering wheel $\delta_{s}$ is taken off of the CAN bus directly. The time-step $\Delta{t}$ is fixed at precisely 10 ms. The wheel base of the vehicle is $L=2.69$m for the 2019 Toyota RAV4. The position and orientation of the ego vehicle in world coordinates are denoted by $\textbf{e}_{t} = (x_{t},y_{t}, \theta_{t})$, and the slip angle with reference to the center of the vehicle is denoted as $\beta$. The full kinematic update at every time-step is produced with:

\begin{align}
\label{bike_eq1}
& x_{t+1} = (v\cos{\theta})\Delta{t} + x_{t} \\
\label{bike_eq2}
& y_{t+1} = (v\sin{\theta})\Delta{t} + y_{t} \\
\label{bike_eq3}
& \theta_{t+1} = (\frac{v\tan{\delta_{w}}\cos{\beta}}{L})\Delta{t} + \theta_{t}
\end{align}
\vskip 0.15in

\subsection{Software-in-the-loop Simulation}
In SIL simulation, the parameters of the vehicle within CARLA are kept mostly in line with what has been tested in CARLA via \textit{Openpilot}’s unofficial simulation tool in order to maintain the highest degree of compatibility possible between the \textit{Openpilot} controls and the CARLA physics control models. Namely, the wheelbase used in simulation is taken from the Tesla Model Y at 2.89m. The weight and steering ratio of the vehicle were changed to further bring the model in line with the Toyota RAV4 at 1530kg and 14.3, respectively. Because the physics model within CARLA determines the dynamics of the vehicle in the SIL simulations, the velocity $\textbf{v}_{t}$, as well as the position and orientation of the vehicle $\textbf{e}_{t}$, are determined by simply querying that information from the CARLA simulator itself at every time instance. Images are passed from CARLA to the \textit{Openpilot} controller in the same manner as our VIL setup. Additional steps are also taken in software to spoof certain CAN messages that \textit{Openpilot} needs to read from the vehicle to operate.

\subsection{Model-in-the-loop Simulation}
In MIL simulation, we follow a pre-defined trajectory $\textbf{e}^{(i)}_{0:T}$ for the vehicle. Here, $i$ is the sample identifier from which the pre-defined trajectory is defined, such that $1 \le i \le n$ for $n$ samples. This notation is ignored unless explicitly required. We follow this predefined trajectory by updating the vehicle's position and orientation across time with a series of speeds $\textbf{v}_{0:T}$ and wheel angles $\delta_{0:T}$, and updating the full position and orientation using \Crefrange{bike_eq1}{bike_eq3}. In this case, there is no vehicle controller in operation. Instead, images from CARLA along the fixed path are passed to \textit{Openpilot}'s \textit{Supercombo} in a standalone manner. This form of simulation allows for "trajectory replay", where experiments performed under VIL or SIL conditions can be repeated in an MIL manner by simply replaying the kinematic response across time, and feeding the resulting images to the perception system. When interpreting the dynamic response of a controlled VIL or SIL experiment, it can be difficult to determine how the perception model contributed to the response at any time $t$ because the trajectory up to that point $\textbf{e}_{0:t}$ has a causal and cascading effect on the images that produce the perception output $\hat{\textbf{e}}_t$, $\hat{d}_t$. Using MIL simulation allows for an analysis of the perception output using a variety of weather and lighting conditions, but controlling for the effect of the vehicle's trajectory.

\section{Experimental Setup}

\begin{figure*}
\centering
\includegraphics[width=\textwidth]{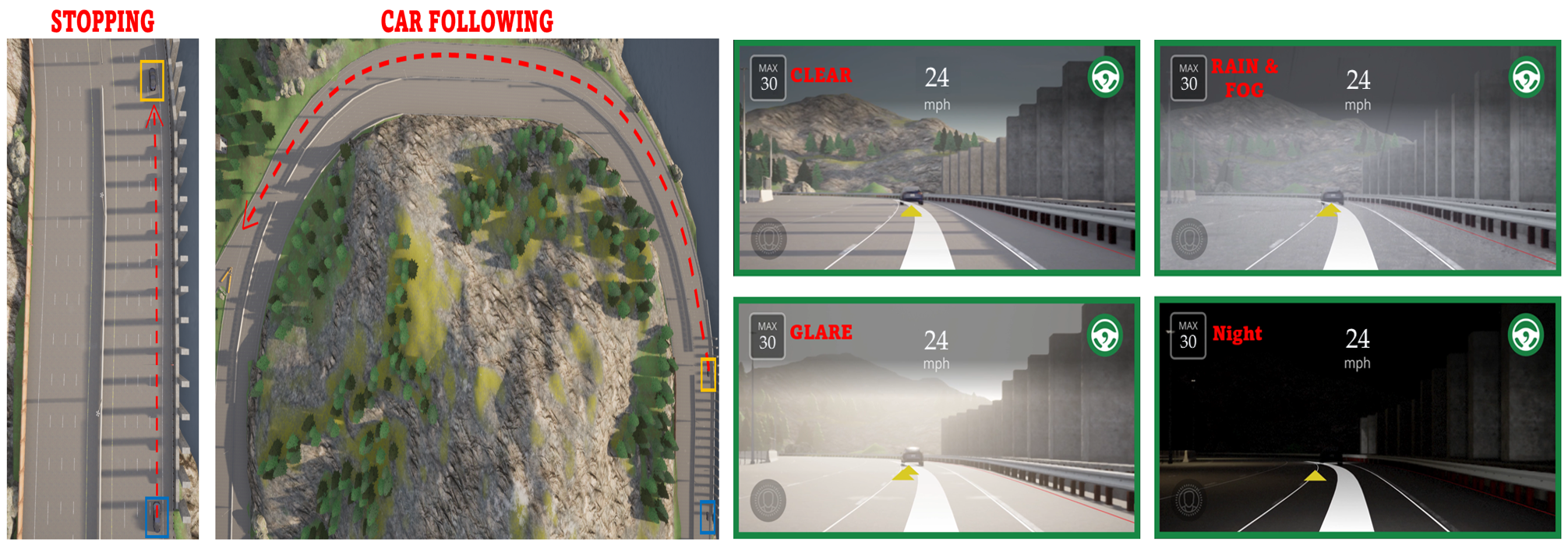}
\caption{Driving types and the primary weather and lighting effects under test are shown. For the two types of driving on the right, the ego vehicle is shown in a blue box, the lead vehicle is shown in an orange box, and the driving path under test is shown as a dashed red line. On the right, an image was captured 8 seconds into the \textbf{Stopping} test for clear weather, rain, sun glare, and night type driving with headlights.
}
\label{driving_types}
\centering
\end{figure*}


In this work, there are three axes of experimentation. The first is simulation type, as described in the previous section. While keeping all other factors identical, we can compare the dynamic responses of SIL and VIL simulations to measure the increase in realism when going from SIL simulation to VIL simulation. The individual components of this difference could be difficult to model separately, with contributions from the response time of vehicle hardware components to the physics of the vehicle itself. Comparing SIL simulations to VIL simulations allows us to begin to gauge that difference in its totality. The second axis is driving type for the ego vehicle. The two types of driving characterized by our experiments are simple car following, and the ability to slow down and stop for a stopped vehicle in front with sufficient spacing between the ego vehicle and the stopped vehicle. These two types of driving capture the essential functions needed in an SAE L2 vehicle. The third and primary axis of experimentation are the factors related to perception: lighting, weather, and lead vehicle type. By modifying elements of the environment that affect perception, we can begin to understand how certain driving conditions might impact driving performance and mitigate against them. Visualizations for each driving type, as well as the primary weather and lighting effects, are shown in \cref{driving_types}. The sample sizes for each experimental combination using the labels defined in this section are shown in \cref{nruns}. The metrics used to measure performance and a quantitative analysis of the results are given in this section. Additionally, the predictive metrics from the perception model are described.

\begin{table}[h]
\begin{tabular}{|c|cc|cc|cc|cc|cc|c|}
\hline
\multirow{2}{*}{\textbf{Driving Type}} & \multicolumn{2}{c|}{\begin{tabular}[c]{@{}c@{}}Baseline\\ (VIL)\end{tabular}} & \multicolumn{2}{c|}{Night} & \multicolumn{2}{c|}{\begin{tabular}[c]{@{}c@{}}Rain\\ Fog\end{tabular}} & \multicolumn{2}{c|}{\begin{tabular}[c]{@{}c@{}}Sun\\ Glare\end{tabular}} & \multicolumn{2}{c|}{\begin{tabular}[c]{@{}c@{}}Sunset\\ Clear\end{tabular}} & \begin{tabular}[c]{@{}c@{}}Sunset\\ Noon\end{tabular} \\ \cline{2-12} 
                                       & S                                     & A                                     & S            & A           & S                                  & A                                  & S                                   & A                                  & S                                    & A                                    & A                                                     \\ \hline
\textbf{Following}                     & 3                                     & 3                                     & 3            & 3           & 3                                  & 3                                  & 3                                   & 3                                  & 3                                    & 3                                    & 0                                                     \\ \cline{1-1}
\textbf{Stopping}                      & 5                                     & 5                                     & 5            & 5           & 5                                  & 5                                  & 5                                   & 5                                  & 5                                    & 5                                    & 5                                                     \\ \hline
\end{tabular}
\centering
\caption{Number of experimental runs $n$ conducted (VIL, SIL, and MIL). "A" and "S" indicate "ambulance lead vehicle" and "sedan lead vehicle" respectively.}
\label{nruns}
\end{table}

\subsection{Driving Types}
Two types of driving were conducted, with the goal of encapsulating the requirements of an SAE L2 vehicle. A bird's eye view of the \textbf{Stopping} and car \textbf{Following} experiments under clear weather conditions are shown in \cref{driving_types}. Due to concerns with preserving experimental hardware, both experiments are conducted at relatively low speeds ($< 40$ mph). The two types of driving under test are:

\begin{itemize}[\leftmargin=0.2in]
\item \textbf{\textit{Stopping}}: This is a test of the obstacle detection system in the SAE L2 controller, as well as the ability to bring the car to a smooth stop with a reasonable amount of distance between the ego vehicle and the leading vehicle. Accounting for lane curvature, a stopped vehicle is placed 125 meters in front of the ego vehicle. The ego vehicle accelerates from 0 mph to a set speed of 30 mph. Data collection begins when \textit{Openpilot} control is activated. The run ends when the vehicle either stops for the lead vehicle, or collides with it.
\item \textbf{\textit{Following}}: This test combines lead vehicle detection with the ability to provide a consistent throttle and brake signal for car following, which is the core expectation of SAE L2 vehicles. \textit{Openpilot} calculates a target following distance based on the relative speed of the ego vehicle and lead vehicle. For the lead vehicle's set speed of 30 mph, this following distance is roughly 22 meters. The ego vehicle accelerates from 0 mph to a set speed of 35 mph. Data collection begins when the ego vehicle comes within 15 meters of the lead vehicle. At that point, the lead vehicle maintains a constant velocity of 30 mph while keeping its current lane by using CARLA's built in autopilot feature, which is a simple PID controller maintaining minimal distance between the vehicle and predefined waypoints in the lane. Accounting for lane curvature, this test lasts for 695m from the starting point of the ego vehicle.
\end{itemize}

\subsection{Weather, Lighting, \& Lead Vehicle}
Weather and lighting effects are applied as a method for applying additional challenges and variance to the \textit{Openpilot}'s perception model. These effects manifest in the lane detection and thus the trajectory prediction, as well as the obstacle detection. Note that all the conditions listed here solely affect perception, and none of these adverse weather conditions effect the dynamic response of the vehicle under test in SIL or VIL simulations.
\begin{itemize}[\leftmargin=0.2in]
\item \textbf{\textit{Baseline}} (VIL): For both types of driving, baseline experiments were performed, which were simply a single human driver performing each experiment on the VIL hardware setup as realistically as possible with no intervention from \textit{Openpilot} or the RAV4's integrated ADAS system. These baseline experiments were performed under clear weather conditions, with the human driver gauging the vehicle response by monitoring the video feed directly from CARLA.
\item \textbf{\textit{SunsetClear}} (VIL, SIL, MIL): Clear weather conditions in the evening.
\item \textbf{\textit{RainFog}} (VIL, SIL, MIL): The hardest possible rain setting within CARLA, mid-afternoon. A small amount of fog is also present (15\% of maximum in CARLA API).
\item \textbf{\textit{SunGlare}} (VIL, SIL, MIL): The sun is positioned to be directly overhead of the lead vehicle for the maximum amount of time for the experiment. This corresponds to a 180 degree azimuth angle in CARLA coordinates for the \textbf{Stopping} experiment, a 270 degree azimuth angle for the \textbf{Following} experiments, and a 10 degree altitude angle for both experiments.
\item \textbf{\textit{Night}} (VIL, SIL, MIL): Clear weather conditions at night, with headlights on for both vehicles.
\item \textbf{\textit{NoonClear}} (VIL, SIL, MIL): Clear weather conditions at midday. This condition was done to add more variance to the results for the ambulance, and was solely done for the \textbf{Stopping} experiment with the ambulance present.
\item \textbf{\textit{FogLevel}} (MIL \textbf{Stopping} Only): The \textbf{\textit{RainFog}} setting is repeated, varying fog percentage from $[5, 100]$ in CARLA to determine how various fog levels in rainy conditions affect perception performance. Experiment consisted of $n=95$ runs, all conducted with the Ambulance lead vehicle.
\item \textbf{\textit{SunAngle}} (MIL \textbf{Stopping} Only): The sun angle is varied for $[0,90]$ degree altitude and $[0, 360]$ degree azimuth angles in CARLA to determine how time of day and sun angle affect perception performance. Experiment consisted of $n=333$ runs, all conducted with the black sedan lead vehicle.
\end{itemize}
The other factor affecting the perception system of \textit{Openpilot} is the choice of the lead vehicle for both types of driving. Headlights were active in night driving experiments for both vehicles. Experiments were conducted with:
\begin{itemize}[\leftmargin=0.2in]
\item \textbf{\textit{Black Sedan}}: A black Tesla model S. Distance  measurement $d_{t}$ is offset 2.39m to establish the position of the rear of the vehicle, as opposed to the center of the vehicle.
\item \textbf{\textit{Ambulance}}: A Ford E-Series Ambulance with full ambulance reflective markings. Distance  measurement $d_{t}$ is offset 3.17m to establish the position of the rear of the vehicle, as opposed to the center of the vehicle.
\end{itemize}

To fully control against the effect of vehicle trajectory, MIL experiments were conducted for all conditions in \cref{nruns} using the vehicle trajectories $\textbf{e}^{(1:n)}_{0:T}$ from the \textbf{VIL : Night : Black Sedan} experiments, and thus kept the same number of samples $n$. The choice of \textbf{Night : Black Sedan} conditions was made arbitrarily. Note that these trajectories were used regardless of the choice of lead vehicle. In the experiments specific to MIL simulation, only the first sample trajectory $\textbf{e}^{(1)}_{0:T}$ from the experiment \textbf{VIL : Night : Black Sedan} was used.

\subsection{Safety, Performance, \& Predictive Metrics}
In both driving types, we focus on safety when measuring performance. The vehicle dynamics should be smooth, the vehicle should stay as close to the center of its own lane as possible, and the vehicle should obviously avoid close calls and collisions while keeping a safe following distance. We also define predictive metrics. These are outputs of our perception model that can be useful to diagnose or predict the vehicle response. The 2D Cartesian position and speed of the ego vehicle are defined as the vector $e_{p}$ and  the scalar $e_{s}$ respectively. The location and speed of the lead vehicle are defined as $l_{p}$ and $l_{s}$ respectively. Notationally, $n$ is reserved for the number of runs for each experiment, shown in \cref{nruns}. We define $n_{e}$ as the number of samples present in each measured trajectory. Our sampling rate for all experiments was 20Hz. Thus, $n_{e} = 20T$ for all runs, while the length of the run in seconds $T$ varied. For convenience, the distance between the two vehicles at any given time instance $t$ is denoted $d_{t} = {\|e_{pt} - l_{pt} \|}$. The metrics used in our analysis are as follows:

\begin{itemize}[\leftmargin=0.2in]
\item \textbf{\textit{Average Centerline Distance}} (\textbf{Following}): Centerline distance is the smallest distance between the line representing the middle of the current lane, and the midpoint of the ego vehicle. In CARLA, only discrete measurements of the lane position are available. Centerline distance is measured as the distance from the ego vehicle (i.e. $e_{p}$) to the line formed by the two nearest lane position measurements denoted by $c_{1}$ and $c_{2}$. Mean centerline distance is defined as:
\vskip 0.05in
\begin{align}
\label{cl_eq}
\CD = \frac{1}{n_{e}} \sum_{t}{\mid \frac{c_{{1}_{t}} - c_{{2}_{t}} \times {e_{{p}_{t}} - c_{{1}_{t}}}}{\|c_{{1}_{t}} - c_{{2}_{t}} \|^2}\mid}
\end{align}
\vskip 0.15in

Note that \cref{cl_eq} can straightforwardly become a predictive metric as well, when the vehicle position $e_{pt}$ is replaced with the model's predicted trajectory for the vehicle $\hat{e}_{pt}$. In the following section when this predictive form is used we will refer to \cref{cl_eq} as $\CDhat$.

\item \textbf{\textit{Minimum time-to-collision}} (TTC) (Stopping): The  time-to-collision measures how long it would take to impact the lead vehicle if the ego vehicle continues at its current speed indefinitely. In this way, TTC can be measured at every time instance. A value very close to $0$ would indicate that a collision occurred. The minimum time to collision is simply the smallest time to collision that occurred in the run, defined as:

\begin{align}
\label{ttc_eq}
\TTC = \min_{t}{\frac{e_{{s}_{t}}}{d_{t}}}
\end{align}
\vskip 0.15in

\item \textbf{\textit{Mean Following Distance}} (\textbf{Following}): \textit{Openpilot} computes a set following distance, given the speed of the ego vehicle and the speed of the lead vehicle. For the \textbf{Following} experiments, this is approximately 22 meters. While this may be higher than what could be considered a safe threshold, we can measure performance relative to how well the vehicle maintained this target distance. Mean following distance is simply defined as:

\begin{align}
\label{fd_eq}
\FD = \frac{1}{n_{e}} \sum_{t}{d_{t}}
\end{align}
\vskip 0.15in

\item \textbf{\textit{Mean Detection Ratio}} (\textbf{Stopping} \& \textbf{Following}): This is a predictive metric that measures the performance of the lead-vehicle detection feature of the perception model. This is simply a measure of what degree the model is \textit{overestimating} the distance between the ego vehicle and the lead vehicle, where a ratio above 1 indicates an overestimate, and a ratio below 1 indicates an underestimate. Where $\hat{d_{t}}$ is the distance estimate from the model at any time $t$, the mean detection ratio is defined as:

\begin{align}
\label{dr_eq}
\DR = \frac{1}{n_{e}} \sum_{t}{\frac{\hat{d}_t}{d_{t}}}
\end{align}
\vskip 0.15in

\end{itemize}

\section{Results}
When analyzing the results of our two experiments through the lens of the metrics outlined in the previous section, we have three main objectives. First, we show that the performance metrics used can be applied to qualitatively illustrate performance. We can see the potential pitfalls of SIL simulation against VIL simulation, and we can also infer at a high level about the implications of varying environmental and situational conditions. Second, we show that it is possible in the \textit{Openpilot} controller to \textit{diagnose} the vehicle's dynamic response. Given that the only variables within each type of experiment are factors that solely effect perception, we show that it is possible to correlate the output of the perception system with safety critical metrics found in the eventual vehicle response. Finally, we take this a step further by showing that it is possible to \textit{predict} the VIL vehicle response with \textit{Openpilot's} perception output in the MIL experiments. This combination allows us to correlate the perception data from our MIL experiments against the highly accurate vehicle response of our VIL experiments, effectively combining the best aspects of both modalities.

\begin{figure}[h]
\captionsetup[subfigure]{oneside,margin={0.65cm,0cm}}
\begin{subfigure}[b]{\textwidth}
    \centering
    \includegraphics[width=0.8\linewidth]{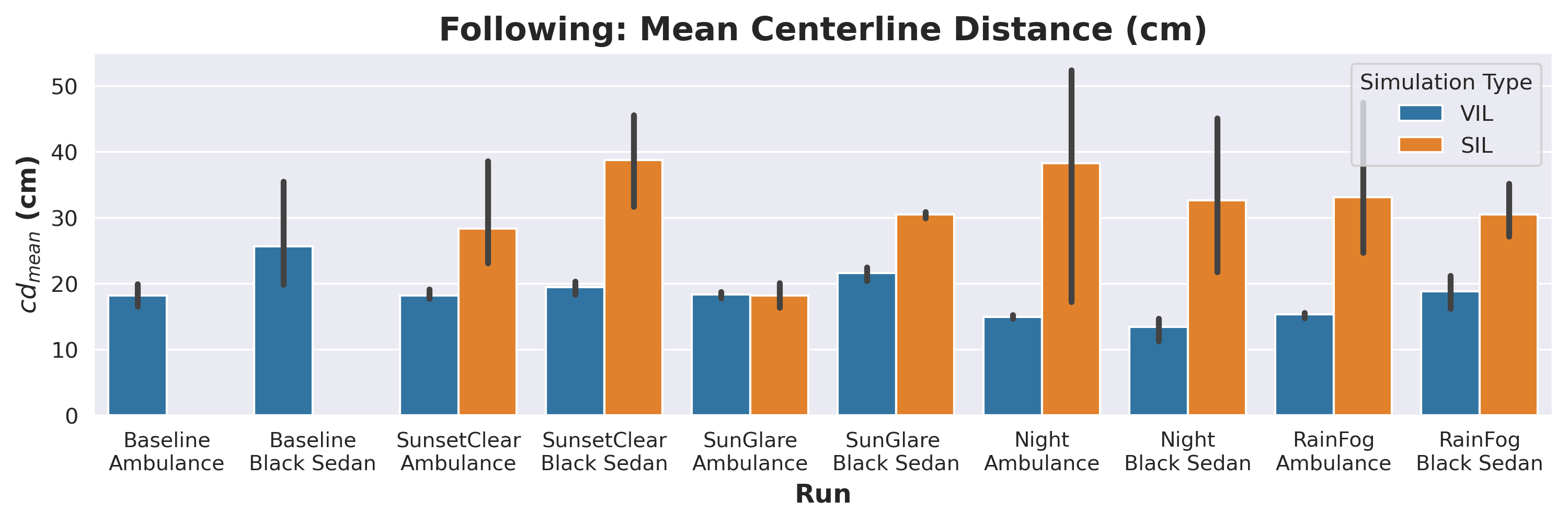}
    \subcaption{$\CD$}
\end{subfigure}%
\hfill
\begin{subfigure}[b]{\textwidth}
    \centering
    \includegraphics[width=0.8\linewidth]{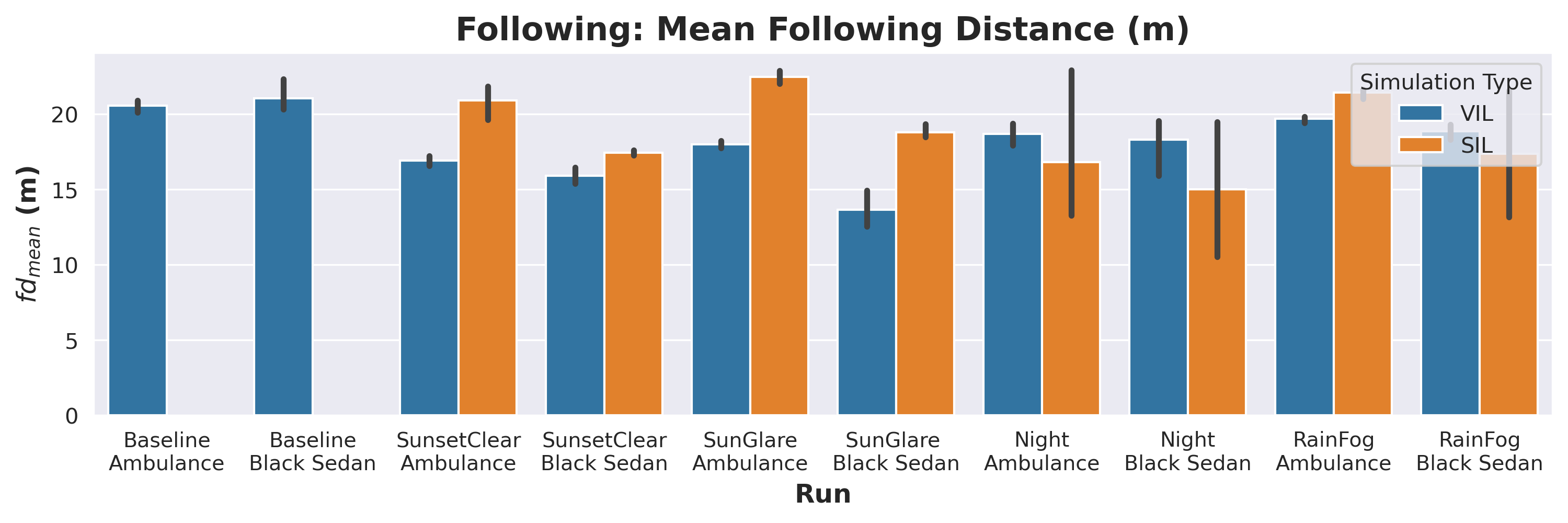}
    \subcaption{$\FD$}
\end{subfigure}%
\hfill
\begin{subfigure}[b]{\textwidth}
    \centering
    \includegraphics[width=0.8\linewidth]{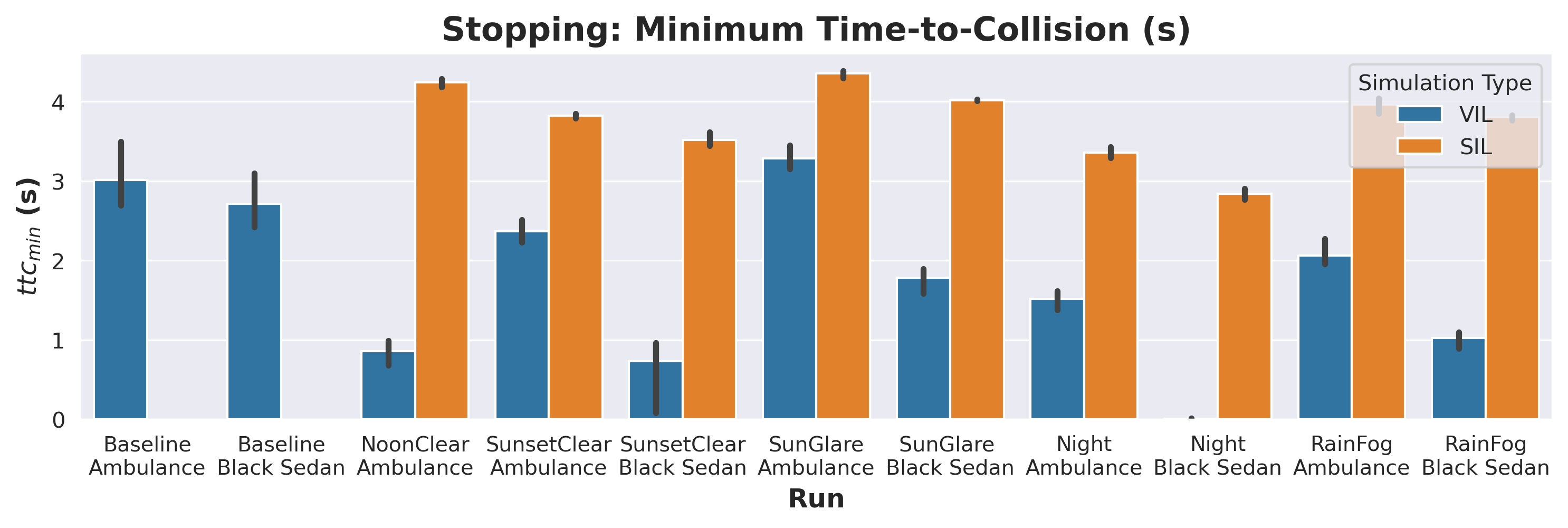}
    \subcaption{$\TTC$}
\end{subfigure}%
    \caption{Performance metrics are shown. Bar length represents the mean score for all runs, with number of runs shown in \cref{nruns}. Vertical black bars indicate the minimum and maximum measured scores for each metric.}
\label{metricbars}
\end{figure}

\subsection{Performance Overview}
The metrics collected are quantitatively shown in \cref{metricbars}. The first comparison to make is the general performance of \textit{Openpilot} against the human driving baseline. In terms of maintaining a low distance from the center of the lane, \textit{Openpilot} running the VIL loop generally meets or outperforms the human driver in the following experiments, both in terms of mean distance, as well as the variance of lane deviation across runs. As might be expected, \textit{Openpilot} is able to control the vehicle with fine tuned consistency and precision in this metric that is measured in centimeters. In terms of the \textbf{Stopping} experiments, we can also observe that \textit{Openpilot} is significantly less conservative in terms of the minimum time to collision than the baseline human driver. The second observation we may point out is that the vehicle in SIL simulation generally had a much more aggressive response to control stimulus than the VIL simulation. This had a mixed bag of effects on performance. While \textbf{Following}, this resulted in more aggressive steering, which resulted in a higher $\CD$ than in the VIL experiments, with more deviation in response as well. More importantly, we see a major difference in the results of the stopping experiments. We notice that the VIL experiments were generally much more dangerous than the SIL experiments. While many of the VIL experiments resulted in collision or near collision, very little can be parsed from the SIL experiments in terms of safety. While the reasoning that explains why some environmental conditions outperformed others is not readily apparent in \cref{metricbars}, the hardware and vehicle implementation on the dynamometer clearly allowed us to produce an experiment that represented a safety critical edge case that would not have been apparent if our testing was limited to SIL simulation. 

We can also begin to examine the effects of weather and lighting at a high level in the case of the VIL simulations. In various instances, the results do not follow intuition. \textbf{SunGlare} produced more centerline deviation and a larger deviation from the target following distance of 22 meters in the car \textbf{Following} experiments. However, \textbf{SunGlare} easily produced the safest $\TTC$ for both the sedan and the ambulance in the \textbf{Stopping} experiments, while environmental deterioration like \textbf{RainFog} outperformed \textbf{SunsetClear} for the black sedan. Indeed, these results both highlight the utility of a VIL simulation platform instead of just a SIL platform, as well as demonstrate the need to investigate the causal effects of the perception system further in order to attribute explanations for these safety critical edge cases.

\begin{figure}[h]
\captionsetup[subfigure]{oneside,margin={0.85cm,0cm}}
\begin{subfigure}[t]{0.4\textwidth}
    \includegraphics[width=\linewidth]{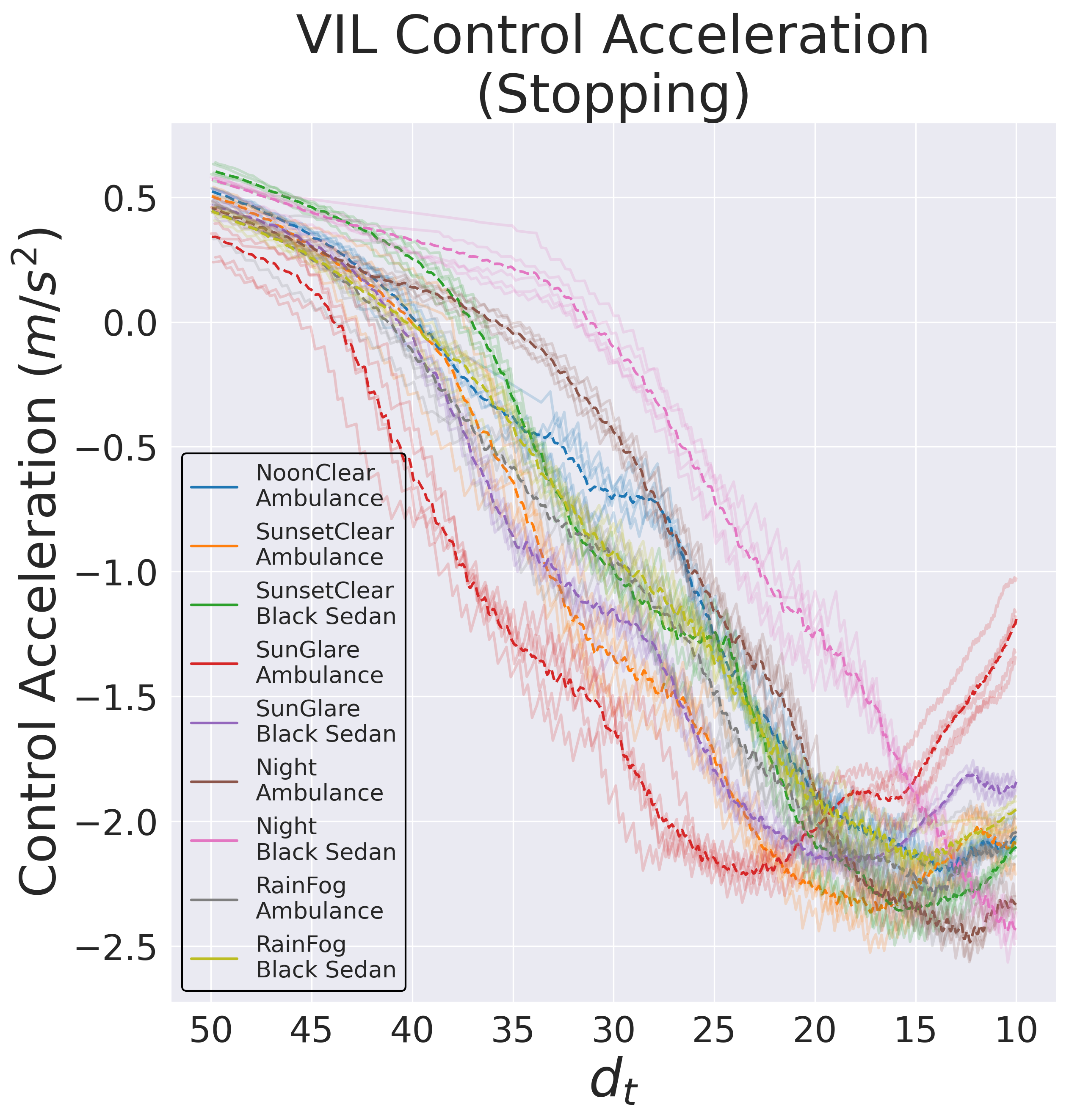}
    \subcaption{$d_{t}$ vs. control accel.}
    \label{rawfig:control-accel}
\end{subfigure}%
\begin{subfigure}[t]{0.4\textwidth}
    \includegraphics[width=\linewidth]{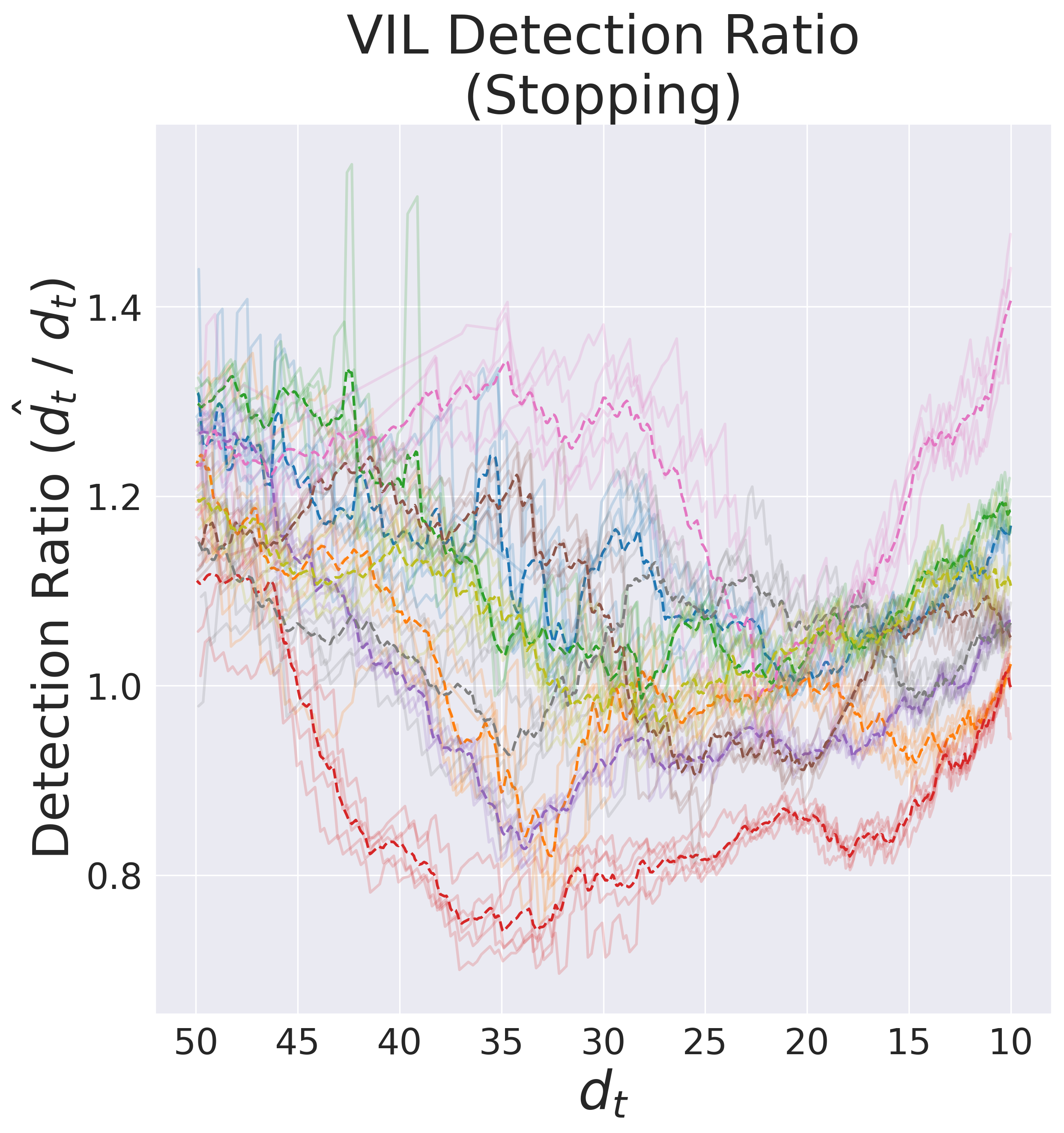}
    \subcaption{$d_{t}$ vs. detection ratio $\hat{d}_{t} / d_{t}$}
    \label{rawfig:obrat}
\end{subfigure}%
\centering
\caption{Raw data for the stopping experiments is shown. \cref{rawfig:control-accel} shows the control acceleration signal supplied to the car during the deceleration curve plotted against lead distance $d_{t}$. \cref{rawfig:obrat} plots the detection ratio for the \textbf{Stopping} experiments in the same distance range. Actual trajectory samples are shown as transparent, while the dotted lines indicate the mean value for each experiment.}
\label{rawfig}
\end{figure}

\subsection{Response Diagnosis}
The results from \cref{metricbars} indicate the various weather, lighting, and choice of lead vehicle imposed in our experiments produced significant deviation in terms of the safety critical response of the vehicle. This is particularly true in the \textbf{Stopping} experiments, which will draw the most of the focus of our analysis. We can begin to understand this response by working backwards from the control commands to the output of the perception system. \textit{Openpilot} controls the longitudinal motion with a single acceleration command given to the vehicle. This acceleration command for each \textbf{Stopping} experiment is shown for $10m < d_{t} < 50m$ in \cref{rawfig:control-accel}. We can see that while the requested acceleration stays mostly consistent in the beginning of each run for all experiments, deviation sets in during deceleration. This deviation correlates as expected with the $\TTC$ metric in \cref{metricbars}. For example, we can clearly see in \cref{rawfig:control-accel} that \textbf{SunGlare Ambulance} had the most aggressive deceleration, while \textbf{Night Black Sedan} had the least aggressive deceleration and resulted in collisions.

Given that all other factors are controlled for in our experiments, we can assume that the causal element affecting the deceleration response time in \cref{rawfig:control-accel} is the scene variation in the images being given to \textit{Supercombo}. We note that this perception system is a deep learning vision model. With this model, an intuitive understanding of why the model performs the way it does may not be possible without access to the model architecture itself. For instance, an understanding of why the model performs safer under sun glare conditions than under normal midday conditions is not available to us. However, we can still use the output of the model to draw direct correlations between how the perception model failed, and the resulting dynamic response. We use the detection ratio $\hat{d}_{t} / d_{t}$ as the baseline performance predictor in the \textbf{Stopping} experiments under the assumption that the differing responses in the \textbf{Stopping} experiments are the result of the perception model producing different estimates $\hat{d}_{t}$ of the lead vehicle distance $d_{t}$. In \cref{rawfig:obrat}, the detection ratio is plotted against lead vehicle distance $d_{t}$. We can see that, like the control acceleration signal of \cref{rawfig:control-accel}, it is a good predictor of $\TTC$. Generally speaking, the more the model \textit{overestimates} the distance of the lead vehicle, the lower the $\TTC$ of the run. To that end, it is clear from \cref{rawfig:obrat} that the choice of lead vehicle had a large impact on this metric, as the position of the black sedan was more consistently overestimated in a comparison against the ambulance.

\begin{figure}[h]
\captionsetup[subfigure]{oneside,margin={0.85cm,0cm}}
\centering
\begin{subfigure}[c]{0.333\textwidth}
    \includegraphics[width=\linewidth]{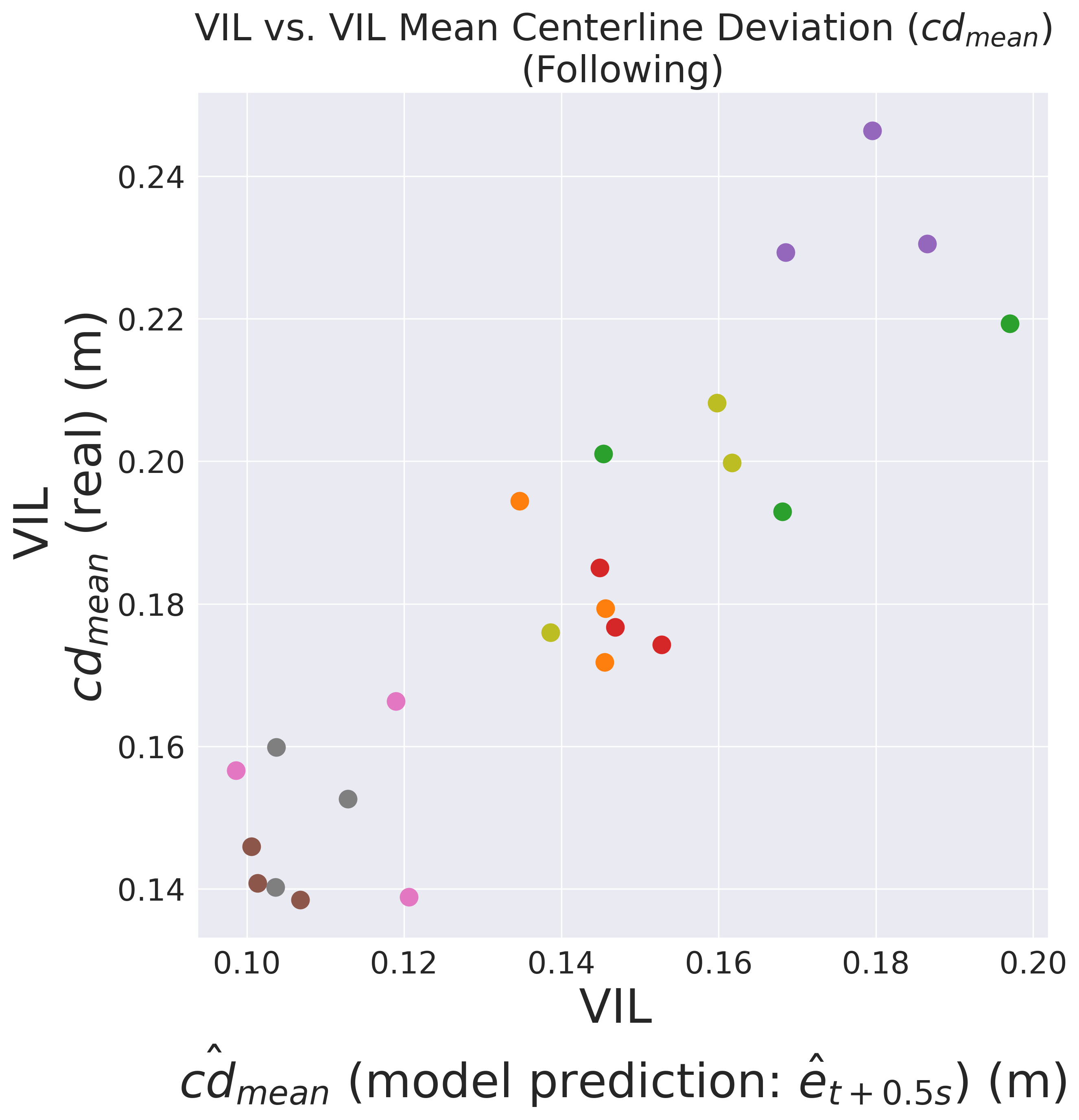}
    \subcaption{$\CDhat$ vs. $\CD$}
    \label{vil2vil:cd}
\end{subfigure}%
\centering
\begin{subfigure}[c]{0.333\textwidth}
    \includegraphics[width=\linewidth]{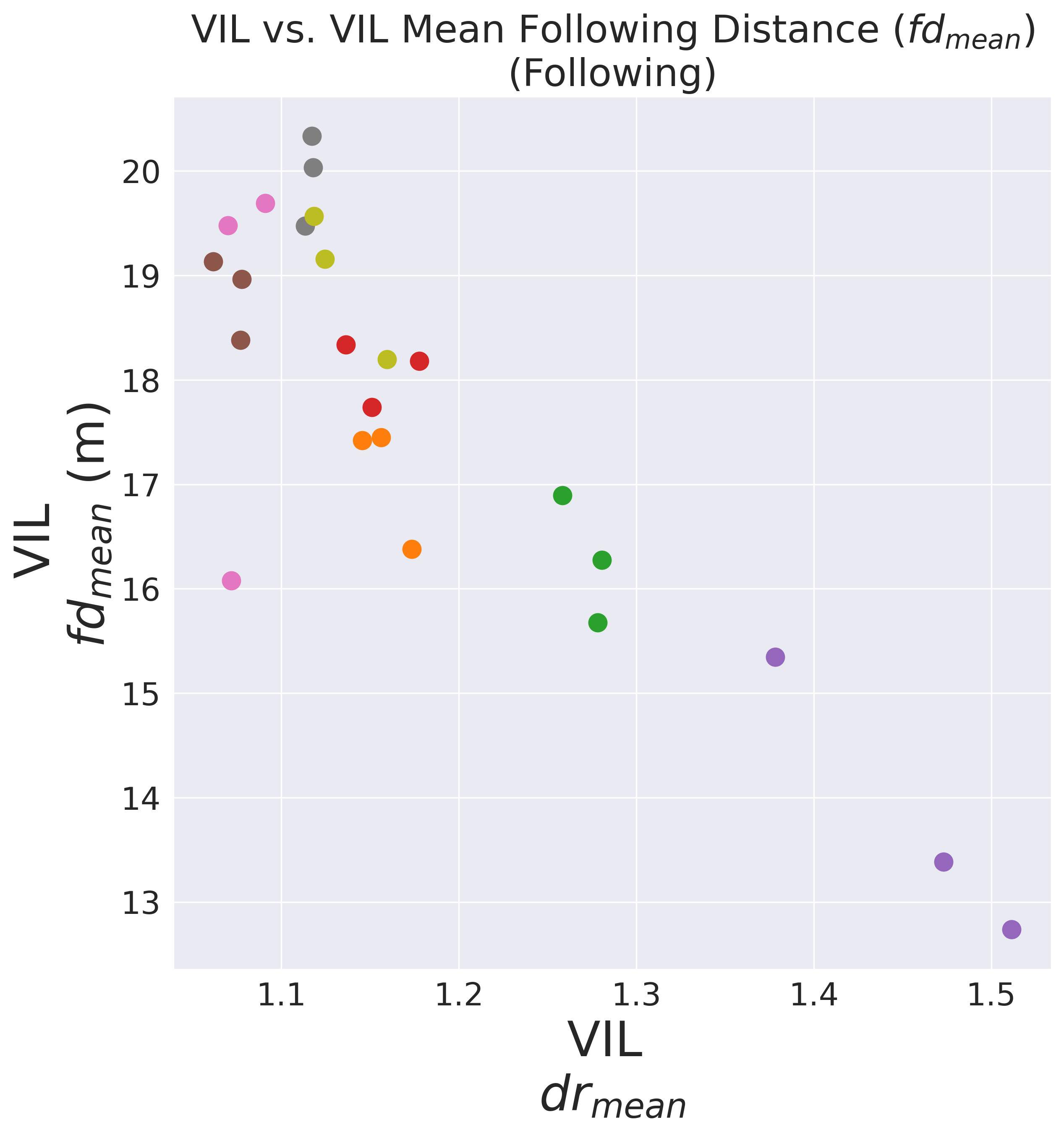}
    \subcaption{$\DR$ vs. $\FD$}
    \label{vil2vil:fd}
\end{subfigure}%
\centering
\begin{subfigure}[c]{0.333\textwidth}
    \includegraphics[width=\linewidth]{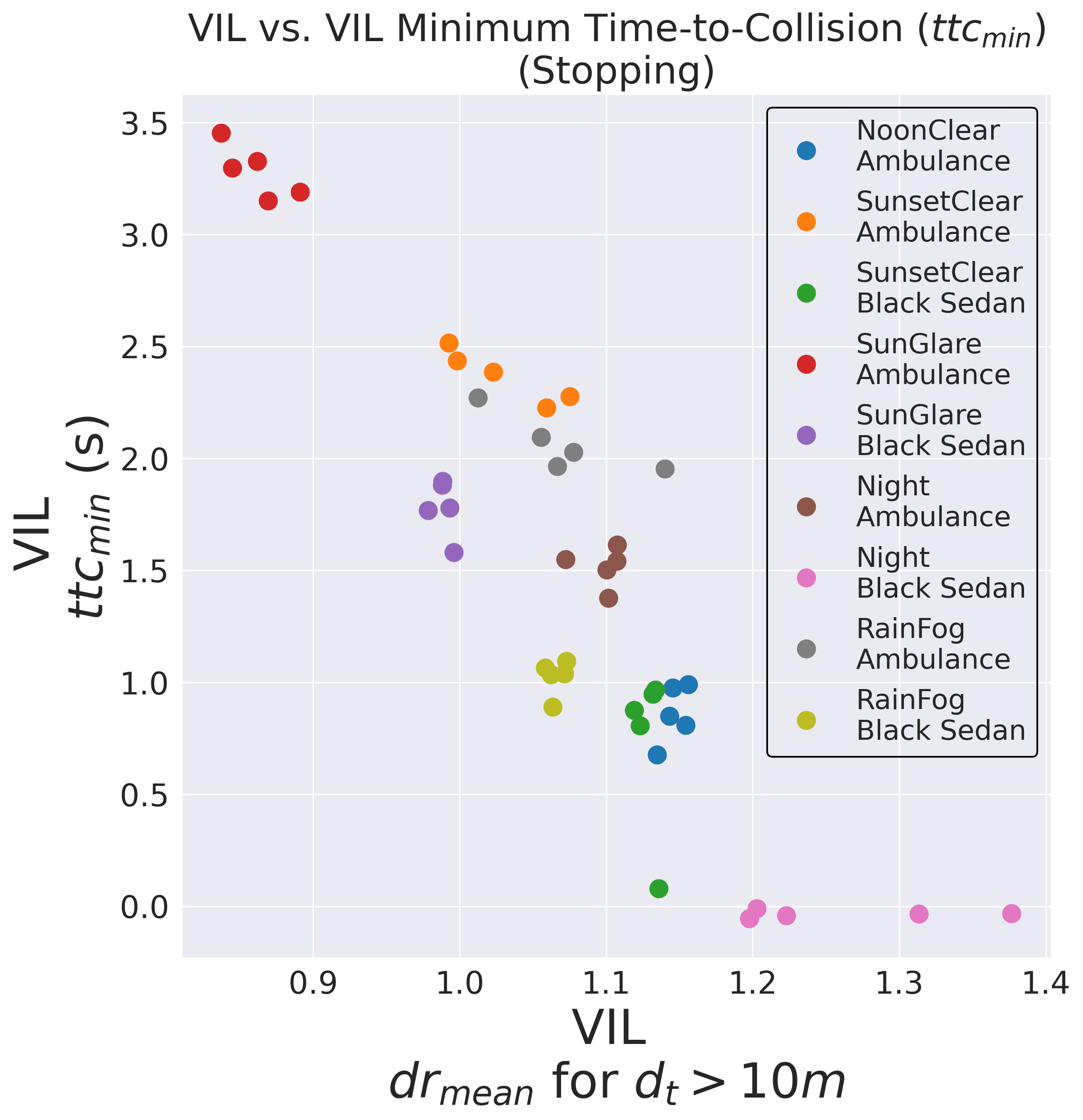}
    \subcaption{$\DR$ vs. $\TTC$}
    \label{vil2vil:ttc}
\end{subfigure}%
    \caption{VIL predictive metrics are plotted against VIL response metrics.}
\label{vil2vil}
\end{figure}

By simply taking the mean detection ratio of the curves in \cref{rawfig:obrat}, we are able to clearly demonstrate the linear relationship between $\TTC$ and $\DR$. Similarly, $\DR$ can be used to predict $\FD$ in the \textbf{Following} experiments, and perhaps most intuitively, the lane deviation of the model $\CDhat$ predicts the resulting lane deviation of the vehicle $\CD$. The linear relationship between these predictive and response metrics for all experiments are shown in \cref{vil2vil}. The causal relationship is visually apparent in all three cases. In \cref{vil2vil:ttc}, we can see that beyond a certain value of $\DR$, a collision becomes imminent. Qualitatively speaking, this effect is the only source of non-linearity in the correlation shown in \cref{vil2vil:ttc}. Pearson Correlation Coefficients (PCC) were produced for all three predictive metrics, shown in the VIL column of \cref{pearsontable}. Correlation is extremely high for all three ($|r| > 0.8$), and all correlations are statistically significant ($p$-value $< 0.05$).

\begin{table}[t]
\centering
\begin{tabular}{|ccc|ccc|ccc|}
\hline
\multirow{2}{*}{\textbf{Driving Type}} & \multirow{2}{*}{\textbf{\begin{tabular}[c]{@{}c@{}}Predictive\\  Metric\end{tabular}}} & \multirow{2}{*}{\textbf{\begin{tabular}[c]{@{}c@{}}Response\\  Metric\end{tabular}}} & \multicolumn{3}{c|}{\textbf{\begin{tabular}[c]{@{}c@{}}VIL-to-VIL\\(diagnosis)\end{tabular}}} & \multicolumn{3}{c|}{\textbf{\begin{tabular}[c]{@{}c@{}}MIL-to-VIL\\(prediction)\end{tabular}}} \\ \cline{4-9} 
                                       &                                                                                        &                                                                                      & $r$                         & $p$-value                      & $n$                      & $r$                         & $p$-value                       & $n$                      \\ \hline
Stopping                               & \begin{tabular}[c]{@{}c@{}}$\DR$\\ ($d_{t} >$ 10m)\end{tabular}                        & $\TTC$                                                                               & -0.89                       & 0.0                            & 45                       & -0.9                        & 0.001                           & 9                        \\ \hline
\rule{0pt}{3ex}Following                              & $\DR$                                                                                  & $\FD$                                                                                & -0.87                       & 0.0                            & 24                       & -0.87                       & 0.02                            & 8                        \\ \hline
\rule{0pt}{3ex}Following                              & $\CDhat$ (+500ms)                                                                   & $\CD$                                                                                & 0.9                         & 0.0                            & 24                       & 0.93                        & 0.001                           & 8                        \\ \hline
\end{tabular}
\caption{Pearson Correlation Coefficient (PCC) Statistical Test Results are shown. Note that the predictive and response metrics for the MIL-to-VIL statistical tests are sample means, defined by \cref{smean}.}
\label{pearsontable}
\end{table}

\subsection{MIL Simulation for Response Prediction}
Given a controller and perception system in which the outputs of these subsystems are transparent, as is the case with \textit{Openpilot}, it becomes possible to gain a rich understanding of the relationship between the perception system and the dynamic response of the vehicle. However, the utility of this understanding is quite limited when searching for new edge-case scenarios if we use VIL simulation alone, given that the performance and safety metrics, which are the metrics we are mostly concerned with in edge-case detection, are already available to use if we commit the resources required for VIL simulation. In the previous section, we diagnosed vehicle behavior using the output of the perception model. Here, we hypothesize that these same predictive metrics from the perception model, when run under certain environmental conditions will still predict the VIL response under those same conditions, even if the trajectory of the vehicle that produced the predictive metrics becomes arbitrary. In other words, we attempt to correlate the predictive metrics of \cref{pearsontable} for MIL simulation against the response metrics of \cref{pearsontable} for VIL simulation. There is no one-to-one relationship between MIL simulation samples and VIL response samples, so we are restricted to correlate these metrics using their respective sample means. For any predictive or response metric $M$, we define the sample mean and the sample min-max pairs respectively as: 
\begin{align}
\label{smean}
& \bar{M} = \frac{1}{n} \sum_{i}{M^{(i)}} \\
& (M^{min},M^{max}) = (\min_{i}{M^{(i)}} , \max_{i}{M^{(i)}})
\label{smax}
\end{align}
\vskip 0.15in
\noindent

\begin{figure}[h]
\captionsetup[subfigure]{oneside,margin={0.85cm,0cm}}
\centering
\begin{subfigure}[c]{0.333\textwidth}
    \includegraphics[width=\linewidth]{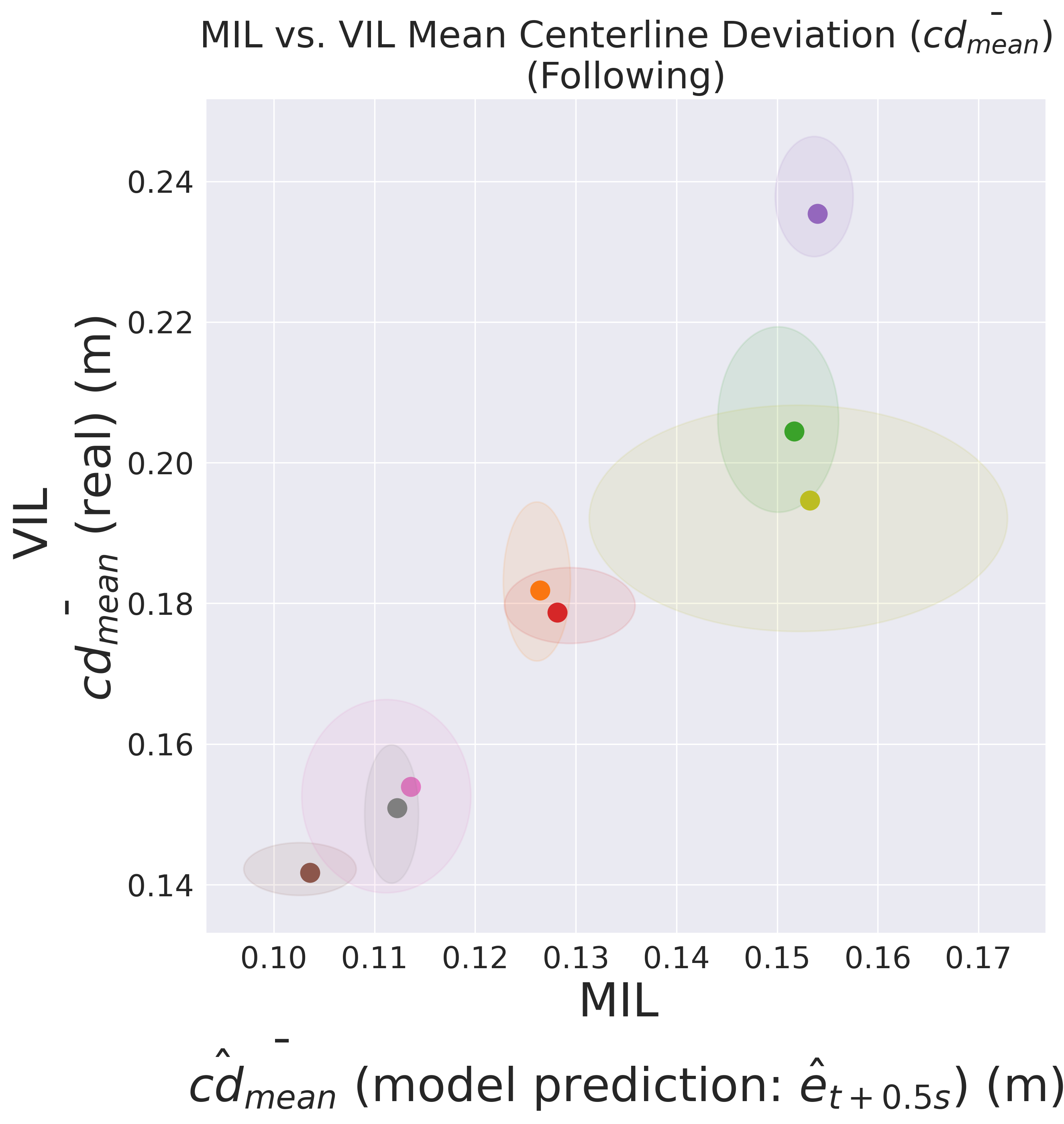}
    \subcaption{$\CDhatbar$ vs. $\bar{\CD}$}
    \label{mil2vil:cd}
\end{subfigure}%
\centering
\begin{subfigure}[c]{0.333\textwidth}
    \includegraphics[width=\linewidth]{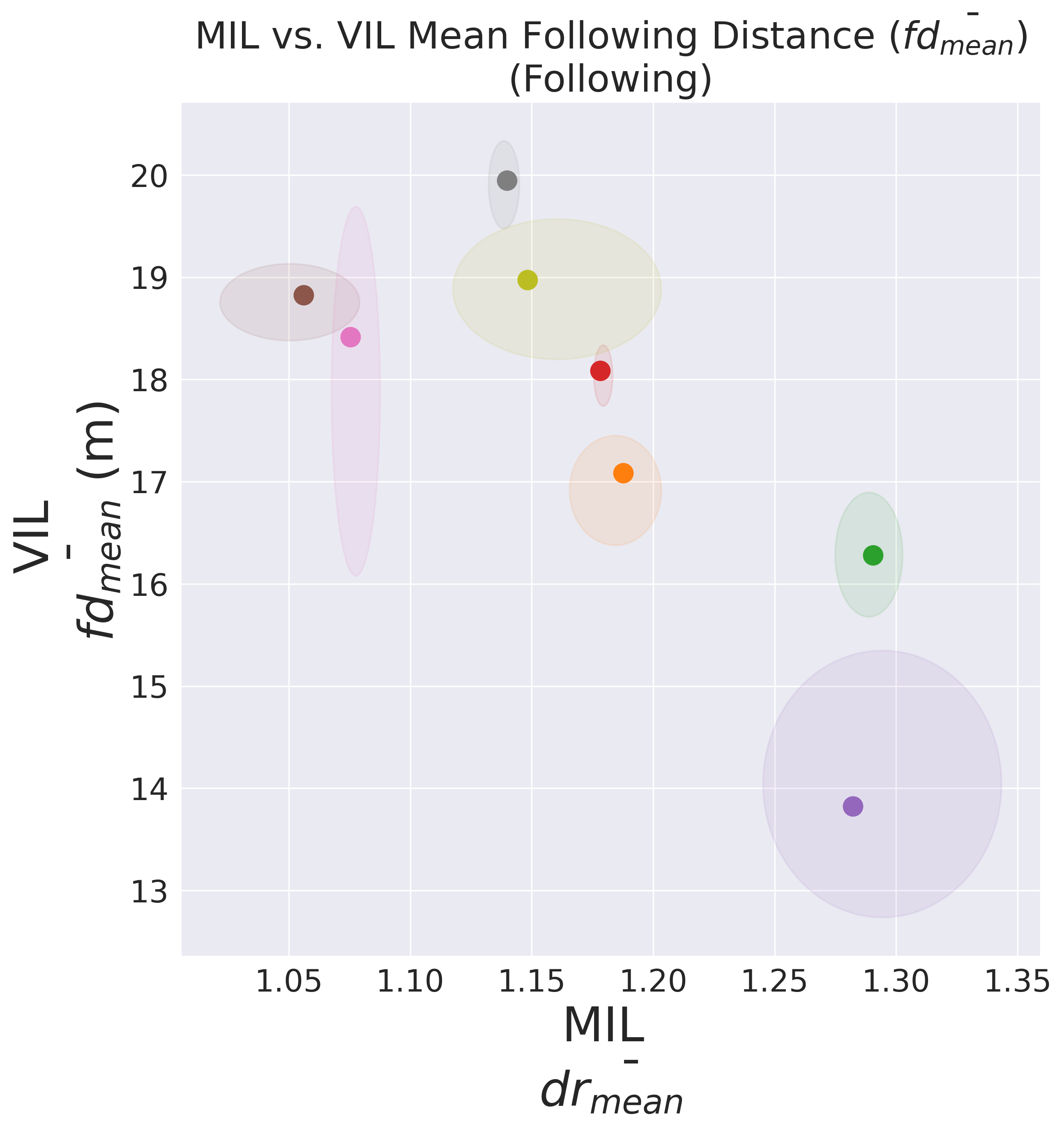}
    \subcaption{$\bar{\DR}$ vs. $\bar{\FD}$}
    \label{mil2vil:fd}
\end{subfigure}%
\centering
\begin{subfigure}[c]{0.333\textwidth}
    \includegraphics[width=\linewidth]{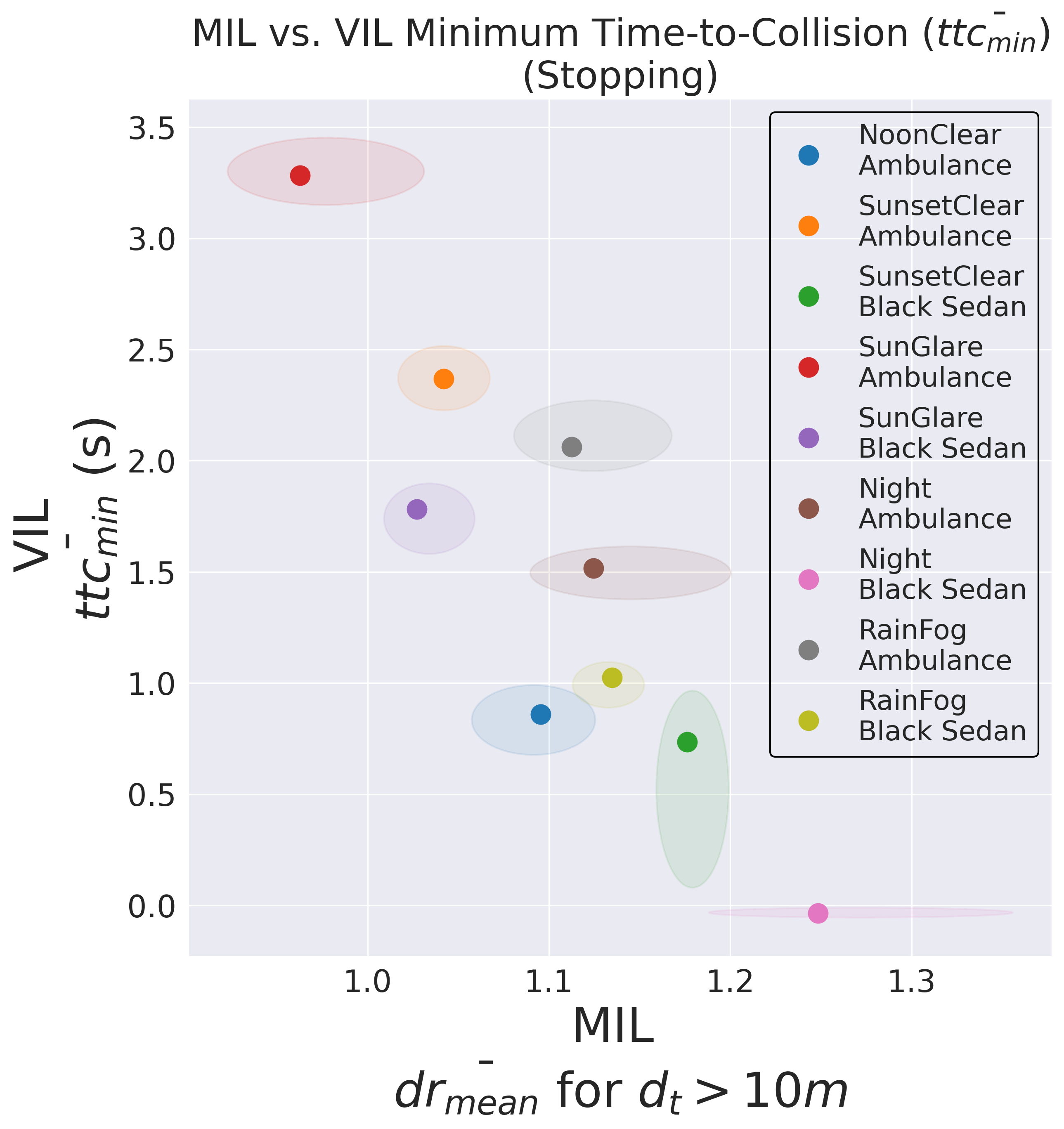}
    \subcaption{$\bar{\DR}$ vs. $\bar{\TTC}$}
    \label{mil2vil:ttc}
\end{subfigure}%
    \caption{MIL predictive metrics are plotted against VIL response metrics. Transparent ovals indicate an estimate of the possible distribution for each experiment, from \cref{smax}.}
\label{mil2vil}
\end{figure}

For all experiments, the predictive sample means for the MIL simulations are plotted against the response sample means of the VIL simulations in \cref{mil2vil}. The sample min-max for these metrics from \cref{smax} define the colored ovals around each plotted sample mean. This is done to provide qualitative context around the potential sampling distribution of both the predictive and response metrics. We can see from \cref{pearsontable} that the the predictive strength from MIL simulation to VIL simulation is still very strong ($|r| > 0.8$) for all of the predictive metrics. $\DR$ remains a strong predictor of both $\TTC$ and $\FD$, while the predicted model trajectory $\CDhat$ is still able to predict the real lane deviation $\CD$, even while the model is being run on a fixed trajectory. While there is a large reduction in sample size by being forced to limit ourselves to regressing against the sample means, the results remain statistically significant ($p$-value $< 0.05$). These results indicate that even when controlling for the cascading impact of the controller and the dynamic response of the vehicle, we can still leverage the perception model in MIL simulation to predict what the response would be under those same environmental conditions in VIL simulation.

\begin{figure}[h]
\captionsetup[subfigure]{oneside,margin={0.85cm,0cm}}
\begin{subfigure}[c]{0.4\textwidth}
    \includegraphics[width=\linewidth]{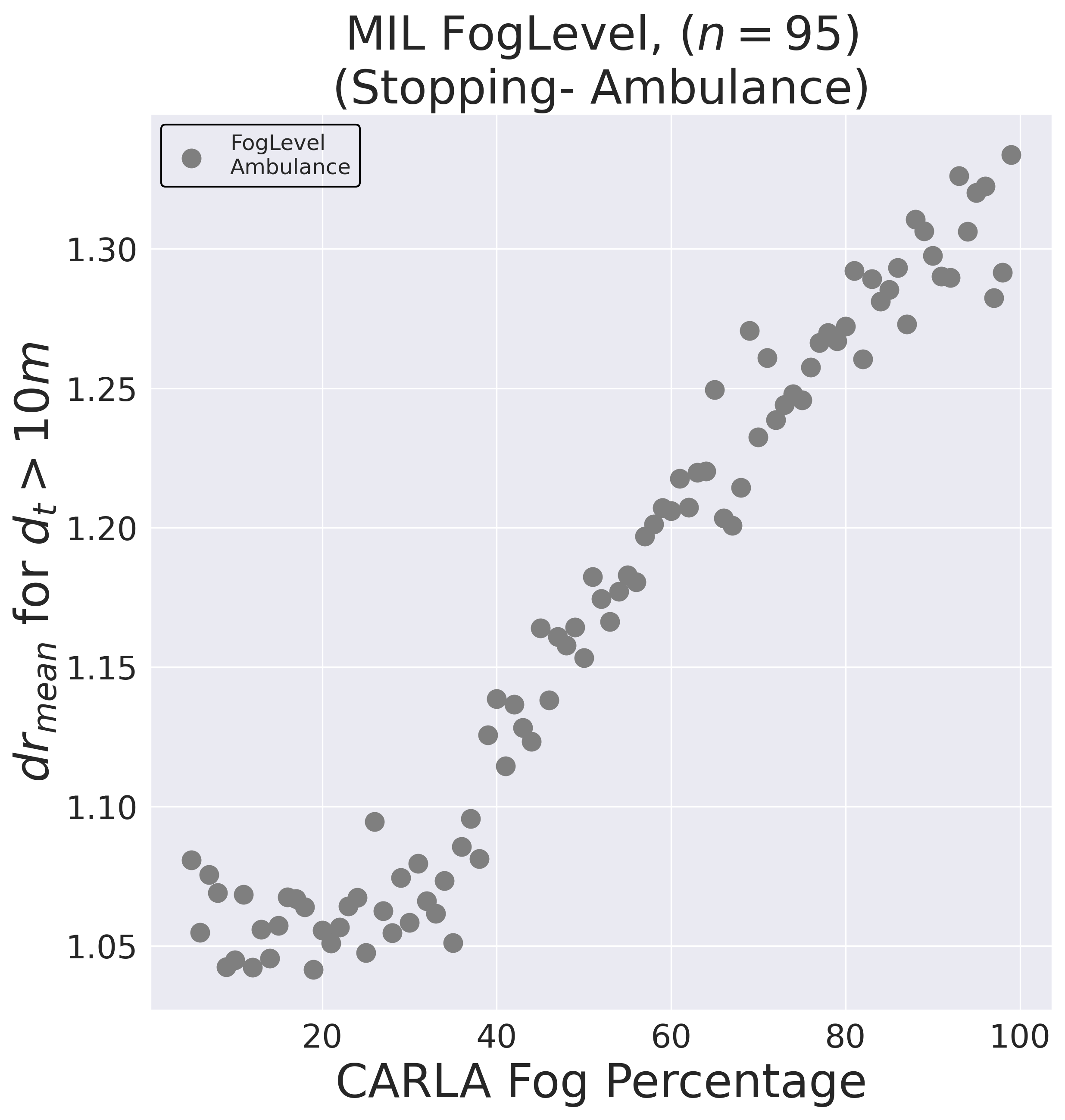}
    \subcaption{FogLevel experiment}
    \label{milfig:fog}
\end{subfigure}%
\centering
\begin{subfigure}[c]{0.4\textwidth}
    \includegraphics[width=\linewidth]{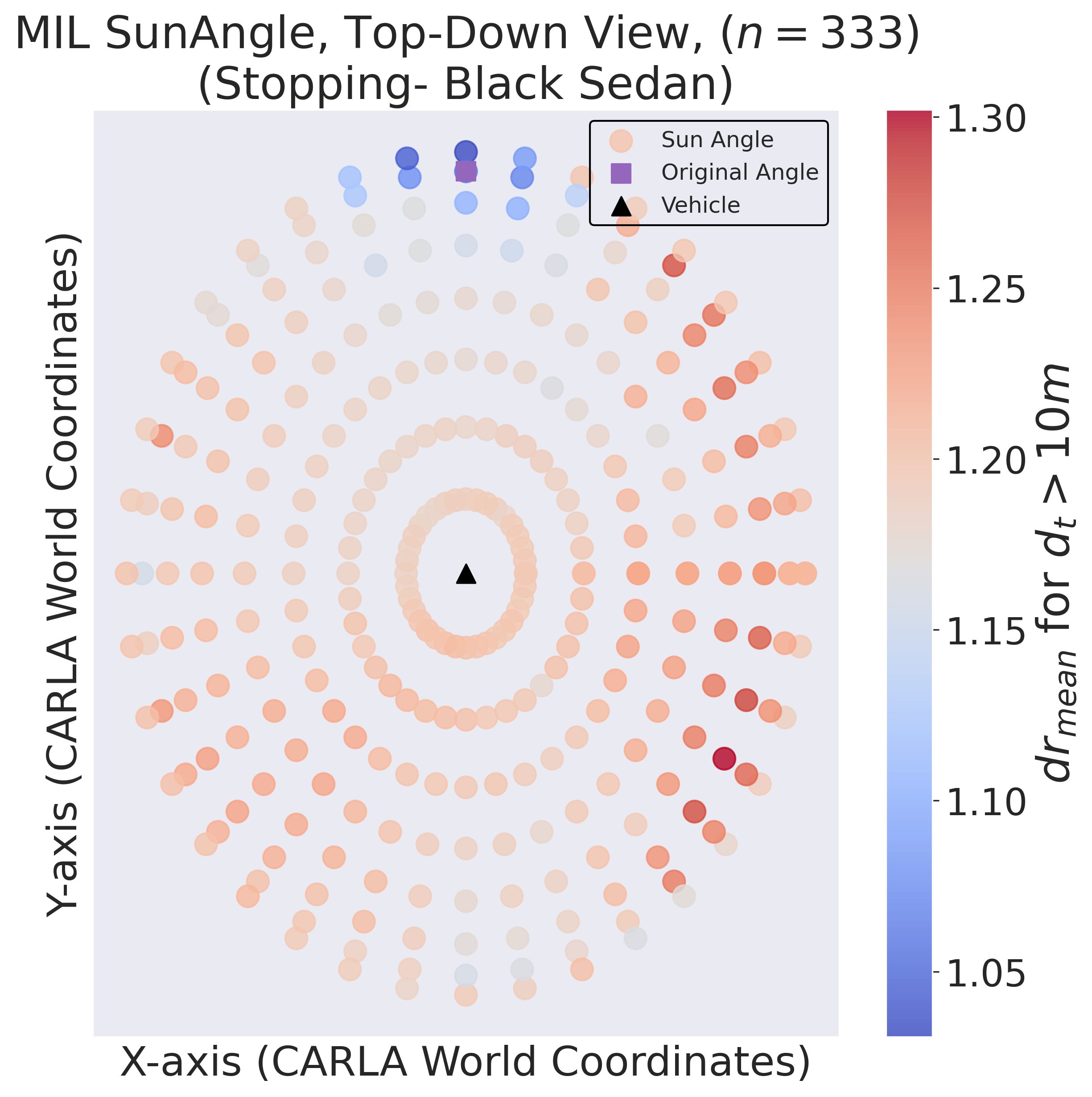}
    \subcaption{SunAngle experiment}
    \label{milfig:sun}
\end{subfigure}%
    \caption{The results for the MIL simulation experiments are shown.}
\label{milfig}
\end{figure}

This combination of using MIL simulation to predict the dynamic response of VIL simulation is a powerful tool for discovering safety critical edge cases. No longer limited by the hardware and real-time nature of VIL simulation, we can perform MIL simulations on any machine capable of running CARLA to gain a deeper and higher resolution understanding of how the vehicle perception system can influence performance. To illustrate this, we once again repeated the \textbf{Stopping} experiments for MIL simulation, but this time with finely tuned changes in environmental conditions and using much larger sample sizes. The results of these experiments are shown in \cref{milfig:fog} and \cref{milfig:sun}. We can see that a richer understanding is visually apparent. In the case of fog level in \cref{milfig:fog}, we can see that under a certain level, fog seems to have little impact on performance. After fog level increases beyond $\approx40\%$, $\DR$ degrades linearly. Comparing with \cref{mil2vil:ttc}, we can estimate that VIL crashes would become imminent ($\TTC = 0$) around a fog level of $>60\%$. In \cref{milfig:sun}, we observe how the various sun angles around the vehicle affect performance when in otherwise clear conditions. We notice that glare directly in front of the vehicle lowers $\DR$ significantly, and is the only sun angle that seems to have this effect. Meanwhile, with the sun on the right-rear of the vehicle, this greatly degrades performance. From the experimental setup, we can deduce that shadows from pillars on the side of the road are having this effect, seen visually in \cref{driving_types}. This angle could be considered a safety critical edge case, discovered only with the combination of VIL and MIL simulation techniques.
\section{Conclusion}
In this work, we leverage the \textit{Openpilot} perception system and controller for SIL, VIL, and MIL simulation of SAE L2 autonomous vehicles. We propose a novel VIL setup, in which images are piped directly from a virtual camera in CARLA to \textit{Openpilot}, which controls the vehicle on a single roller dynamometer with steering enabled synchronized wheel speed (SeSwS). Our VIL setup is compared to an SIL setup against safety and performance metrics, and the performance of the VIL response is diagnosed with respect to the output of the perception model. Additionally, a combination approach to simulation is introduced, where the real dynamic response of VIL simulation is predicted from a MIL simulation strategy. The real-world accuracy of VIL response, paired with the speed and simplicity of MIL simulation, allow for a deep exploration of realistic safety-edge cases for autonomous vehicles, with minimized cost. This combinational approach is used to further explore safety edge cases for our system around the effects of fog and sun angle.

\section*{Declarations}

\section{Abbreviations}
\begin{itemize}
\item CAN- Controller area network
\item AI - Artificial Intelligence
\item SIL- Software-in-the-loop
\item VIL- Vehicle-in-the-loop
\item MIL- Model-in-the-loop 
\end{itemize}

\section{Ethical Approval and Consent to Participate}
Not Applicable

\section{Consent for Publication}
Not Applicable

\section{Availability of data and material}
The data produced in this work is available from the corresponding author on request, including raw dynamics data and video from experimentation.

\section{Competing Interests}
The authors declare that they have no competing interests.

\section{Funding}
The authors would like to acknowledge the support from the Collaborative Sciences Center for Road Safety for funding this work.

\section{Author Contribution}
The authors confirm contribution to the paper as follows: study concept and algorithm design: J. Beck, S. Chakraborty; data acquisition: J. Beck, S. Huff; VIL/hardware setup S. Huff, J. Beck; analysis and interpretation of results: J. Beck, S. Chakraborty; draft manuscript preparation: J. Beck, S. Huff, S. Chakraborty. All authors reviewed the results and approved the final version of the manuscript.

\section{Acknowledgements}



\bibliographystyle{ieeetr}
\bibliography{Ref_CH,Ref_CHF,Ref_NW,Ref_SC,Ref_SS,Ref_MET}

\begin{thebibliography}{10}

\bibitem{carlini2017towards}
N.~Carlini and D.~Wagner, ``Towards evaluating the robustness of neural networks,'' in {\em 2017 IEEE Symposium on Security and Privacy (SP)}, pp.~39--57, IEEE Computer Society, 2017.

\bibitem{madrytowards}
A.~Madry, A.~Makelov, L.~Schmidt, D.~Tsipras, and A.~Vladu, ``Towards deep learning models resistant to adversarial attacks,'' in {\em International Conference on Learning Representations}, 2022.

\bibitem{miller2021accuracy}
J.~P. Miller, R.~Taori, A.~Raghunathan, S.~Sagawa, P.~W. Koh, V.~Shankar, P.~Liang, Y.~Carmon, and L.~Schmidt, ``Accuracy on the line: on the strong correlation between out-of-distribution and in-distribution generalization,'' in {\em International Conference on Machine Learning}, pp.~7721--7735, 2021.

\bibitem{feuer2022meta}
B.~Feuer, A.~Joshi, and C.~Hegde, ``A meta-analysis of distributionally-robust models,'' {\em arXiv preprint arXiv:2206.07565}, 2022.

\bibitem{amani2021safe}
S.~Amani, C.~Thrampoulidis, and L.~Yang, ``Safe reinforcement learning with linear function approximation,'' in {\em Proceedings of the 38th International Conference on Machine Learning} (M.~Meila and T.~Zhang, eds.), vol.~139 of {\em Proceedings of Machine Learning Research}, pp.~243--253, PMLR, 18--24 Jul 2021.

\bibitem{dalal2018safe}
G.~Dalal, K.~Dvijotham, M.~Vecerik, T.~Hester, C.~Paduraru, and Y.~Tassa, ``Safe exploration in continuous action spaces,'' {\em arXiv preprint arXiv:1801.08757}, 2018.

\bibitem{srinivasan2020learning}
K.~Srinivasan, B.~Eysenbach, S.~Ha, J.~Tan, and C.~Finn, ``Learning to be safe: Deep rl with a safety critic,'' {\em arXiv preprint arXiv:2010.14603}, 2020.

\bibitem{Yang2023}
Q.~Yang, T.~D. Sim{\~a}o, S.~H. Tindemans, and M.~T.~J. Spaan, ``Safety-constrained reinforcement learning with a distributional safety critic,'' {\em Machine Learning}, vol.~112, pp.~859--887, Mar 2023.

\bibitem{thananjeyan2021recovery}
B.~Thananjeyan, A.~Balakrishna, S.~Nair, M.~Luo, K.~Srinivasan, M.~Hwang, J.~E. Gonzalez, J.~Ibarz, C.~Finn, and K.~Goldberg, ``Recovery rl: Safe reinforcement learning with learned recovery zones,'' {\em IEEE Robotics and Automation Letters}, vol.~6, no.~3, pp.~4915--4922, 2021.

\bibitem{emam2022safe}
Y.~Emam, G.~Notomista, P.~Glotfelter, Z.~Kira, and M.~Egerstedt, ``Safe reinforcement learning using robust control barrier functions,'' {\em IEEE Robotics and Automation Letters}, pp.~1--8, 2022.

\bibitem{cheng2019endtoend}
R.~Cheng, G.~Orosz, R.~M. Murray, and J.~W. Burdick, ``End-to-end safe reinforcement learning through barrier functions for safety-critical continuous control tasks,'' {\em Proceedings of the AAAI Conference on Artificial Intelligence}, vol.~33, pp.~3387--3395, Jul. 2019.

\bibitem{mindom2021assessing}
P.~S.~N. Mindom, A.~Nikanjam, F.~Khomh, and J.~Mullins, ``On assessing the safety of reinforcement learning algorithms using formal methods,'' in {\em 2021 IEEE 21st International Conference on Software Quality, Reliability and Security (QRS)}, pp.~260--269, IEEE, 2021.

\bibitem{Fulton_Platzer_2018}
N.~Fulton and A.~Platzer, ``Safe reinforcement learning via formal methods: Toward safe control through proof and learning,'' {\em Proceedings of the AAAI Conference on Artificial Intelligence}, vol.~32, Apr. 2018.

\bibitem{Alshiekh:2018:Safe}
M.~Alshiekh, R.~Bloem, R.~Ehlers, B.~Könighofer, S.~Niekum, and U.~Topcu, ``Safe reinforcement learning via shielding,'' {\em Proceedings of the AAAI Conference on Artificial Intelligence}, vol.~32, Apr. 2018.

\bibitem{li2020formal}
X.~Li, {\em A formal methods approach to interpretability, safety and composability for reinforcement learning}.
\newblock PhD thesis, Boston University, 2020.

\bibitem{eryilmaz2014novel}
U.~Eryilmaz, H.~S. Tokmak, K.~Cagiltay, V.~Isler, and N.~O. Eryilmaz, ``A novel classification method for driving simulators based on existing flight simulator classification standards,'' {\em Transportation research part C: emerging technologies}, vol.~42, pp.~132--146, 2014.

\bibitem{chen2001nads}
L.~Chen, Y.~Papelis, G.~Waston, and D.~Solis, ``Nads at the university of iowa: A tool for driving safety research,'' in {\em Proceedings of the 1st human-centered transportation simulation conference}, 2001.

\bibitem{dosovitskiyCARLAOpenUrban2017b}
A.~Dosovitskiy, G.~Ros, F.~Codevilla, A.~Lopez, and V.~Koltun, ``{{CARLA}}: {{An Open Urban Driving Simulator}},'' in {\em Proceedings of the 1st {{Annual Conference}} on {{Robot Learning}}}, pp.~1--16, {PMLR}, Oct. 2017.

\bibitem{rongLGSVLSimulatorHigh2020a}
G.~Rong, B.~H. Shin, H.~Tabatabaee, Q.~Lu, S.~Lemke, M.~Možeiko, E.~Boise, G.~Uhm, M.~Gerow, S.~Mehta, E.~Agafonov, T.~H. Kim, E.~Sterner, K.~Ushiroda, M.~Reyes, D.~Zelenkovsky, and S.~Kim, ``{{LGSVL Simulator}}: {{A High Fidelity Simulator}} for {{Autonomous Driving}},'' in {\em 2020 {{IEEE}} 23rd {{International Conference}} on {{Intelligent Transportation Systems}} ({{ITSC}})}, pp.~1--6, 2020.

\bibitem{ramakrishnaANTICARLAAdversarialTesting2022}
S.~Ramakrishna, B.~Luo, C.~B. Kuhn, G.~Karsai, and A.~Dubey, ``{{ANTI-CARLA}}: {{An Adversarial Testing Framework}} for {{Autonomous Vehicles}} in {{CARLA}},'' in {\em 2022 {{IEEE}} 25th {{International Conference}} on {{Intelligent Transportation Systems}} ({{ITSC}})}, pp.~2620--2627, 2020.

\bibitem{BeckAutomated2022}
J.~Beck, R.~Arvin, S.~Lee, A.~Khattak, and S.~Chakraborty, ``Automated vehicle data pipeline for accident reconstruction: New insights from lidar, camera, and radar data,'' {\em Accident Analysis and Prevention}, vol.~180, p.~106923, 2023.

\bibitem{FremontScenic}
D.~J. Fremont, T.~Dreossi, S.~Ghosh, X.~Yue, A.~L. Sangiovanni-Vincentelli, and S.~A. Seshia, ``Scenic: A language for scenario specification and scene generation,'' in {\em Proceedings of the 40th ACM SIGPLAN Conference on Programming Language Design and Implementation}, PLDI 2019, (New York, NY, USA), p.~63–78, Association for Computing Machinery, 2019.

\bibitem{comma.aiOpenpilotOpenSource}
{comma.ai}, ``Openpilot \textemdash{} open source advanced driver assistance system.'' https://www.comma.ai/openpilot, 2023.

\end{thebibliography}

\end{document}